\documentclass{article}

\usepackage{microtype}
\usepackage{graphicx}
\usepackage{subfigure}
\usepackage{siunitx}
\usepackage{booktabs} % for professional tables
\usepackage{bm}
\usepackage{amsmath}
\usepackage{multirow}
\usepackage{todonotes}
\usepackage{enumitem}
\usepackage{natbib}
\usepackage{subfigure}
\usepackage{float}
\usepackage{graphicx}
\usepackage[utf8]{inputenc} % allow utf-8 input
\usepackage[T1]{fontenc}    % use 8-bit T1 fonts
\usepackage{hyperref}       % hyperlinks
\usepackage{url}            % simple URL typesetting
\usepackage{booktabs}       % professional-quality tables
\usepackage{amsfonts}       % blackboard math symbols
\usepackage{nicefrac}       % compact symbols for 1/2, etc.
\usepackage{microtype}      % microtypography
\usepackage[symbol]{footmisc}

\usepackage{hyperref}

\usepackage[accepted]{icml2020}

\icmltitlerunning{Revisiting One-vs-All Classifiers for Predictive Uncertainty and Out-of-Distribution Detection in Neural Networks}

\newlength{\sfigwidth}
\graphicspath{{figures/}} %Setting the graphicspath

 \usepackage[compact]{titlesec}
 \titlespacing{\section}{0pt}{1ex}{0ex}
 \titlespacing{\subsection}{0pt}{1ex}{0ex}
 \titlespacing{\subsubsection}{0pt}{0.5ex}{0ex}
  \newcommand{\reducespacebetweenfigureandcaption}{\vspace{-0.5em}} % nips
\begin{document}

\cfoot{\thepage}
\setlength{\footskip}{3em}
\vspace{-2em}

\twocolumn[
\icmltitle{Revisiting One-vs-All Classifiers for Predictive Uncertainty and Out-of-Distribution Detection in Neural Networks}

\begin{icmlauthorlist}
\icmlauthor{Shreyas Padhy}{goo,aires}
\icmlauthor{Zachary Nado}{goo}
\icmlauthor{Jie Ren}{goo}
\icmlauthor{Jeremiah Liu}{goo}
\icmlauthor{Jasper Snoek}{goo}
\icmlauthor{Balaji Lakshminarayanan}{goo}
\end{icmlauthorlist}

\icmlaffiliation{aires}{Work done as an AI Resident}
\icmlaffiliation{goo}{Google Research, Brain Team}

\icmlcorrespondingauthor{Shreyas Padhy}{shreyaspadhy@google.com}

\icmlkeywords{Machine Learning, ICML}

\vskip 0.3in
]

\printAffiliationsAndNotice{}

\begin{abstract}
Accurate estimation of predictive uncertainty in modern neural networks is critical to achieve well calibrated predictions and detect out-of-distribution (OOD) inputs. The most promising approaches have been predominantly focused on improving model uncertainty (e.g. deep ensembles and Bayesian neural networks) and post-processing techniques for OOD detection (e.g. ODIN and Mahalanobis distance). However, there has been relatively little investigation into how the parametrization of the probabilities in discriminative classifiers affects the uncertainty estimates, and the dominant method, softmax cross-entropy, results in misleadingly high confidences on OOD data and under covariate shift. We investigate alternative ways of formulating probabilities using (1) a one-vs-all formulation to capture the notion of “none of the above”, and (2) a distance-based logit representation to encode uncertainty as a function of distance to the training manifold. We show that one-vs-all formulations can improve calibration on image classification tasks, while matching the predictive performance of softmax without incurring any additional training or test-time complexity.
\end{abstract}

\section{Introduction}
\label{introduction}

In recent years, the calibration of deep learning models used for predictive discrimination tasks has become very important, especially in the domains of healthcare \citep{esteva2017dermatologist, dusenberry2020analyzing}, self-driving vehicles \citep{bojarski2016end}, and
more generally, AI safety \citep{amodei2016concrete}. 
Estimating the \textit{epistemic} uncertainty is especially valuable, as it captures the model's lack of knowledge about data. Recent work in the field has dealt with three different paradigms of measuring predictive uncertainty, namely (1) in-distribution calibration of models measured on an i.i.d.~test set, (2) robustness under dataset shift, and (3) anomaly/out-of-distribution (OOD) detection, which measures the ability of models to assign low confidence predictions on inputs far away from the training data. 
Unfortunately, it has been recognized that vanilla predictive models fail to provide accurate quantification of predictive uncertainty \citep{guo2017calibration, hein2019relu}.
In recent years, many methods have been proposed to tackle this, including deep ensembles \citep{lakshminarayanan2017simple}, Bayesian methods \citep{mackay1992bayesian, neal2012bayesian, blundell2015weight}, and post-processing methods such as temperature scaling \citep{guo2017calibration}, ODIN \citep{liang2017enhancing}, and Mahalanobis distance \citep{lee2018simple}. 
However, the best performing methods under dataset shift either require additional memory and time complexity, or are hard to scale up for large datasets and architectures \citep{snoek2019can}. 

In this work, we first study the contribution of the loss function used during training to the quality of predictive uncertainty of models. Specifically, we show why the parametrization of the probabilities underlying softmax cross-entropy loss are ill-suited for uncertainty estimation.
We then propose two simple replacements to the parametrization of the probability distribution: (1) a one-vs-all normalization scheme that does not force all points in the input space to map to one of the classes, thereby naturally capturing the notion of “none of the above”, and (2) a distance-based logit representation to encode uncertainty as a function of distance to the training manifold.

\section{Choice of parametrization of probabilities}
In a supervised learning setting, we assume we have access to training data  $\left\{\left(\bm{x}_{i}, y_{i}\right)\right\}_{i=1}^{N}$ drawn from a training distribution $p_{\texttt{in}}(\bm{x}, y)$, where $\bm{x}$ denotes the feature vector and $y\in\{1,\ldots,K\}$ denotes the class label. We train a model to make predictions on unlabeled testing data $\left\{\bm{x}_{j}\right\}_{j=1}^{M}$ by parametrizing the embeddings $f_{\bm{\theta}}(\bm{x})$ of a neural network model into a predictive probability distribution $p(\hat{y}|\bm{x})$, and minimizing the empirical loss function calculated over a mini-batch. \\ 

In the traditional discriminative setting, the logit outputs of a neural network are calculated from the latent space embeddings through an affine transformation $\bm{z}_j = \bm{w}_j^T f_{\bm{\theta}} (\bm{x}) + b_j$. 
The probability distribution is then calculated through the softmax normalization, 
\begin{equation}
    p_{\texttt{S}}(\hat{y}^{(k)} | \bm{x}) = \frac{\exp \left(\boldsymbol{w}_{k}^{\top} f_{\bm{\theta}}(\boldsymbol{x})+b_{k}\right)}{\sum_{j=1}^K \exp \left(\boldsymbol{w}_{j}^{\top} f_{\bm{\theta}}(\boldsymbol{x})+b_{j}\right)}. 
\end{equation}
The parameters $\bm{\theta}$ are learned by minimizing \emph{cross-entropy} loss, which corresponds to maximizing  the log-likelihood:
\begin{equation}
    \mathcal{L}_{\texttt{S}}\left(\hat{y}^{(k)} | \bm{x}\right) =- \log \left( p_S(\hat{y}^{(k)} | \bm{x})\right).
\end{equation}
%\item 

This parametrization of the probabilities that are output by the model has been shown to result in misleadingly high-confidence predictions under dataset shift, and on OOD data \citep{guo2017calibration, hendrycks2016baseline}. Firstly, generating the class logits through an affine transformation of the embedding space has been shown to result in a confidence landscape that is invariant to the distance from the in-distribution embeddings in the model's latent space \citep{hein2019relu}. Secondly, the softmax normalization function explicitly maps the class logits into a $K$-dimensional simplex, for a $K$ class problem, and this closed-world assumption of the input space results in highly confident, yet mis-classified predictions on entirely OOD points.

\subsection{Distance-based logits}
A natural way of mitigating the first issue is to explicitly encode a notion of distance in the formulation of the class logits from the learnt embeddings of a predictive model. This can be achieved by scaling the logits to be proportional to a function of distance of a point from the training manifold, and there has been previous work that enforces this parametrization in different ways \citep{zadeh2018deep, xing2019distance}. A particularly simple approach is one suggested by \citet{macedoisotropic} for OOD detection, where the logits $\bm{z}_j$ are defined as the negative Euclidean distance between the embeddings and the weights of the last dense layer, $\bm{z}_j = - \lVert f_{\bm{\theta}}(\bm{x}) - \bm{w}_j \rVert$. The probability distribution is calculated similarly with a softmax distribution, given as
\begin{equation}
			\label{eq:dm_loss}
    p_{\texttt{D}}(\hat{y}^{(k)} | \bm{x}) = \frac{\exp \left(- \lVert f_{\bm{\theta}}(\bm{x}) - \bm{w}_j \rVert \right)}{\sum_{j} \exp \left(- \lVert f_{\bm{\theta}}(\bm{x}) - \bm{w}_j \rVert \right)}
\end{equation}
and the loss function, called \emph{Distinction Maximization (DM) or Isotropic Maximization} loss is calculated by maximizing the log-likelihood, given as
\begin{equation}
    \mathcal{L}_{\texttt{D}}\left(\hat{y}^{(k)} | \bm{x}\right) =- \log \left( p_D(\hat{y}^{(k)} | \bm{x})\right)
\end{equation}
According to \citet{macedoisotropic}, the weights of the last layer $\bm{w}_j$ end up approximately learning the class prototypes of the $j$th class during optimization in a mini-batch setting. 

\begin{figure}[h]
    \centering
    \includegraphics[width=0.45\textwidth]{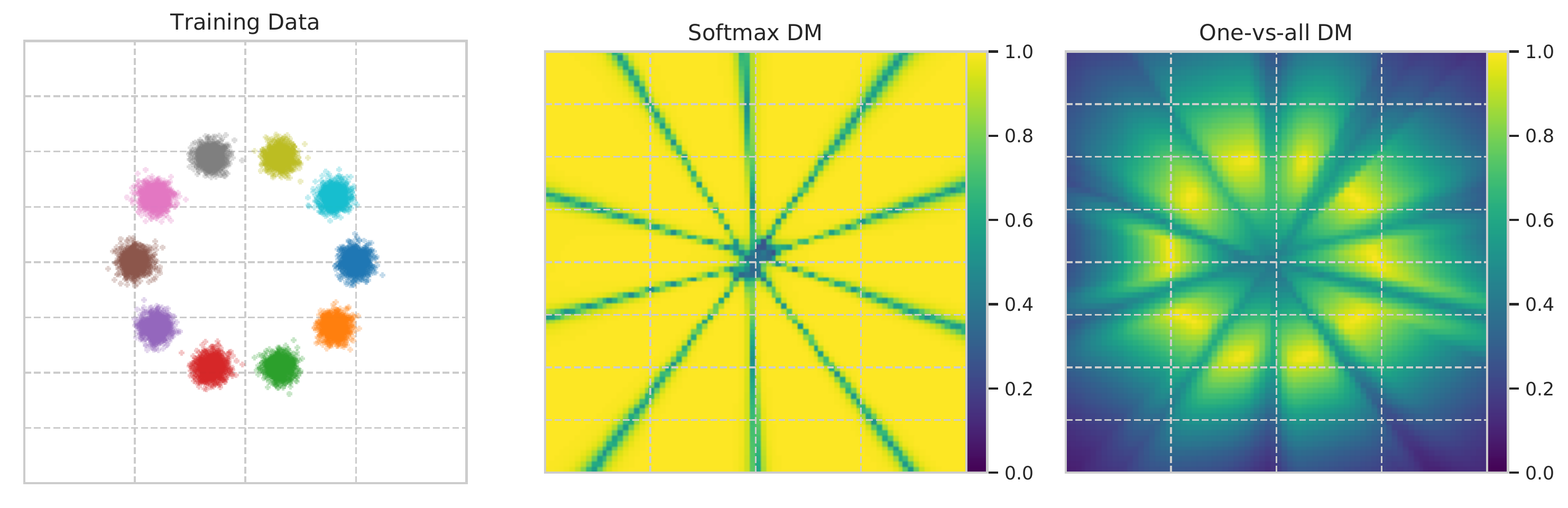}
    \caption{Confidence landscapes for a 2-dimensional toy example with 10 classes \textbf{(left)}. We train a 2 layer fully-connected neural network with the different loss functions (further details in Appendix \ref{subsection:2d}). It can be seen that a distance-based loss function with the softmax normalization results in infinite regions of space with uniform confidence divided by the decision boundaries \textbf{(center)}, whereas a distance-based method with one-vs-all normalization learns to predict higher confidences centered on the training manifold \textbf{(right)}.}
    \label{fig:confidence-softmax-cross-entropy}
\end{figure}

In practice, DM loss achieves better performance on OOD detection compared to the baseline of softmax cross-entropy. However, we observe that for in-distribution calibration and under dataset shift, a distance-based loss function that uses the softmax normalization performs worse, due to either a systematic under-confidence issue (which is very apparent on the CIFAR-10 dataset in particular, see Figure~\ref{fig:rel_cifar10}), or due to the underlying assumption of the softmax normalization that maps the domain to one of the $K$ classes. Due to the relative normalization of the softmax function, even small differences in the distance of a point between class centers are highly exacerbated due to the exponentiation of the logits, resulting in points far away from the training manifold being predicted with a uniform confidence. In summary, the softmax normalization undoes the distance constraint imposed in this specific formulation, and we demonstrate this pathology in Figure~\ref{fig:confidence-softmax-cross-entropy} for a 2-dimensional toy example. 

\subsection{Revisiting Neural One-vs-All Classifiers}
In order to avoid imposing the closed-world assumption of the softmax normalization, we explore an alternate formulation of the predictive discrimination task, where the $K$ class problem is split into $K$ independent binary tasks, each parametrized by a sigmoid normalization. More concretely, for affine-transformed logits $\bm{z}_j = \bm{w}_j^{\top}f_{\bm{\theta}}(\bm{x}) + b_j$, instead of a softmax normalization, we take $K$ independent sigmoids, so that
\begin{equation}
    p_{\texttt{OVA}}(\hat{y}^{(k)} | \bm{x}) = \frac{1}{1 + \exp \left(-( \bm{w}_k^{\top}f_{\bm{\theta}}(\bm{x}) + b_k )\right)}
\end{equation}
whereas for a distance-based logit given by $\bm{z}_j = - \lVert f_{\bm{\theta}}(\bm{x}) - \bm{w}_j \rVert$, we have
\begin{equation}
			\label{eq:ova_dm}
    p_{\texttt{OVADM}}(\hat{y}^{(k)} | \bm{x}) = \frac{2}{1 + \exp \left(- \lVert f_{\bm{\theta}}(\bm{x}) - \bm{w}_k \rVert \right)}.
\end{equation}
The factor of 2 here is a technical implementation detail that allows us to map the logits in a distance-based formulation from the $(-\infty, 0]$ domain to the $[0, 1]$ domain. Since the logits are defined as the negative of the Euclidean distances, they are constrained to be in the negative domain, and the maxima of the sigmoid function in this range is 0.5 (at a value of 0), which corresponds to a distance of 0 to the learned class center. The factor of 2 therefore maps a distance of 0 to the learned class center to a confidence of 1. 

In both cases, the loss function for a single example is then given by a sum over the $K$ independent binary probabilities, maximizing the binary log-likelihood for the positive class and minimizing it for the negative classes

\begin{align}
   \mathcal{L}_{\texttt{OVA}} = -\log p(\hat{y}^{(k)} | \bm{x}) - \sum_{j=1, j\neq k}^{K}   
     \log \left(1 - p(\hat{y}^{(j)} | \bm{x})\right).
\label{eq:ova_loss}
\end{align}

One-vs-all formulations of loss functions have been studied before, especially in the context of multi-class classification \citep{duan2003multi, rifkin2004defense} and for Support Vector Machines (SVMs) \citep{liu2005one, anthony2007image, mathur2008multiclass}. More recently, one-vs-all formulations have been used to train classifiers for open-set recognition \citep{shu2017doc}, due to their ability to encode a "none of the above" class naturally. The predictive uncertainty of one-vs-all classifiers were very recently explored by \citet{franchi2020one}, where individual deep ensemble members were trained to specialize in independent one-vs-all tasks for a multiclass problem. 

Formulating the loss function as in Equation~(\ref{eq:ova_loss}) offers a few benefits, however.
Firstly, by replacing the softmax normalization by $K$ independent sigmoids, models are able to predict a higher confidence centered on the training manifold for each specific class versus the rest of the training manifold (see Figure \ref{fig:confidence-softmax-cross-entropy}).
Secondly, by minimizing a linear combination of the binary log-likelihoods, we can train a single deterministic model to learn to optimize over the K binary tasks jointly, avoiding additional overhead during training or evaluation. 
Thirdly, a distance-based one-vs-all formulation learns more meaningful class-centers in a minibatch fashion, compared to DM loss with a softmax normalization. We demonstrate this behavior for the 2-dimensional toy example in Figure~\ref{fig:classcenters}, where it can be seen that softmax DM loss does not learn good representations of the centers of embeddings. Instead, due to the relative nature of the softmax normalization, it is more likely to learn a pivot point in embedding space, where if all points of a class are closest to their pivot point instead of the others, they will still get correctly classified. One-vs-all DM loss does not suffer from this issue, as the formulation in Equation~\ref{eq:ova_dm} maps a distance of 0 to a class center in the embedding space, to a confidence score of exactly 1, whereas this is not strictly true for a softmax-based formulation.

\section{Experimental Results}
We report the performance of our proposed methods on a variety of different tasks that measure in-distribution calibration, robustness under covariate shift, and performance on OOD data. The quality of uncertainty for a predictive distribution is calculated using \textbf{Expected Calibration Error} (ECE)~\citep{guo2017calibration}, which is a binned measure of the difference between the accuracy of a model and its predicted confidence, where the confidence is defined as the maximum predicted class probability for that example. For points with binned confidences $B_i$, if we define $\operatorname{acc}(B_i)$ as the accuracy and $\operatorname{conf}(B_i)$ as the confidence, the ECE is then defined as
$\text{ECE} = \sum_{i}{\frac{B_i}{N}|\operatorname{acc}\left(B_i\right) - \operatorname{conf}\left(B_i\right)|}$. We use $N=15$ bins to calculate ECE for our experiments.

\subsection{Image Classification Tasks}
\begin{figure}[ht]
    \centering
    \begin{subfigure}[Expected Calibration Error]{
          \includegraphics[width=0.9\linewidth]{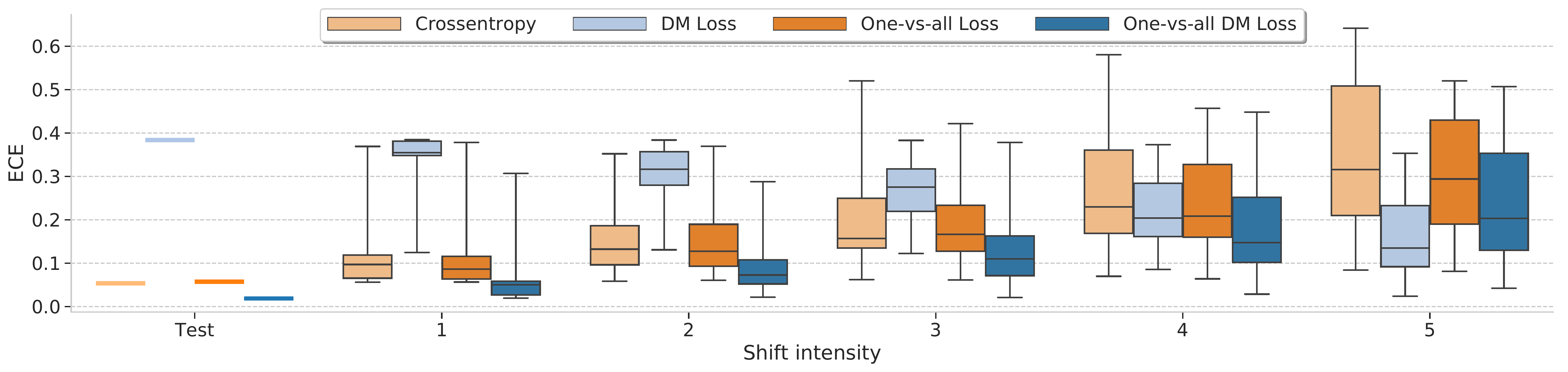}
   } \end{subfigure}
    \begin{subfigure}[Accuracy]{
          \includegraphics[width=0.9\linewidth]{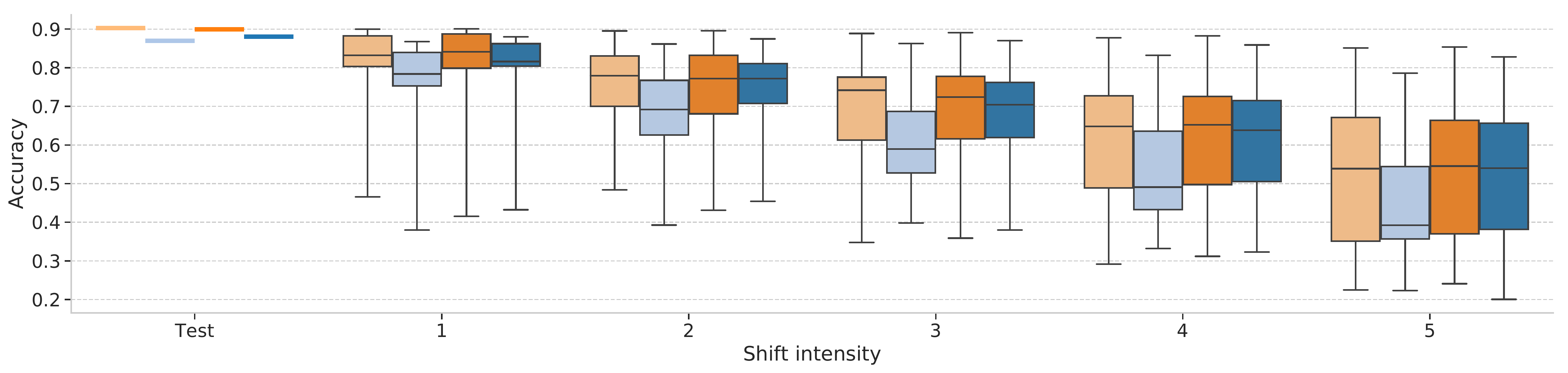}
   } \end{subfigure}
\reducespacebetweenfigureandcaption
\caption{
Expected Calibration Error and Accuracy under dataset shift for CIFAR-10-C. The box plots show the median, quartiles, minimum, and maximum performance per method. One-vs-all classifiers maintain similar predictive accuracy under covariate shift as the softmax cross-entropy baseline, while consistently achieving better calibration under covariate shift, especially as shift increases, and adding a distance-based logit representation further improves ECE performance. Data at \href{https://tensorboard.dev/experiment/pot4YujvReuNEmzSoUmOYw/\#scalars&_smoothingWeight=0}{this url}.
}
\label{fig:accuracy-cifar10c}
\end{figure}

To evaluate the behavior of discriminative models trained under our proposed loss functions, we benchmark the methods using the methodology first proposed in \citep{snoek2019can} on the CIFAR-10-C and CIFAR-100-C datasets. These image datasets contain corrupted versions of the original test datasets, with 19 types of corruption applied with 5 levels of intensity each \citep{hendrycks2019benchmarking}. In our experiments, models are trained with identical architectures and optimization setups, and for loss functions with a distance-based formulation, the last layer of the model is replaced with a distance-based layer, as defined in Equation~\eqref{eq:dm_loss} and Equation~\eqref{eq:ova_dm}. For CIFAR-10, we train a ResNet 20 model using the different loss functions (further implementation details are given in Appendix \ref{subsection:cifar10}), where the distance-based layer weights are initialized by zeros as specified in \citep{macedoisotropic}, though we notice no difference if they are initialized using the standard random initialization in TensorFlow \citep{abadi2016tensorflow}. From Figure \ref{fig:accuracy-cifar10c}, we can see that one-vs-all methods consistently perform better under covariate shift while maintaining predictive accuracy compared to the baseline of softmax crossentropy, with a distance-based one-vs-all formulation performing the best amongst the methods. A distance-based softmax formulation suffers from consistently under-confident predictions, resulting in ECE improving as prediction accuracy drops at higher levels of shift, and this behavior is further discussed in Appendix~\ref{section:reliability}.

\begin{figure}[ht]
    \centering
    \begin{subfigure}[Expected Calibration Error]{
          \includegraphics[width=0.9\linewidth]{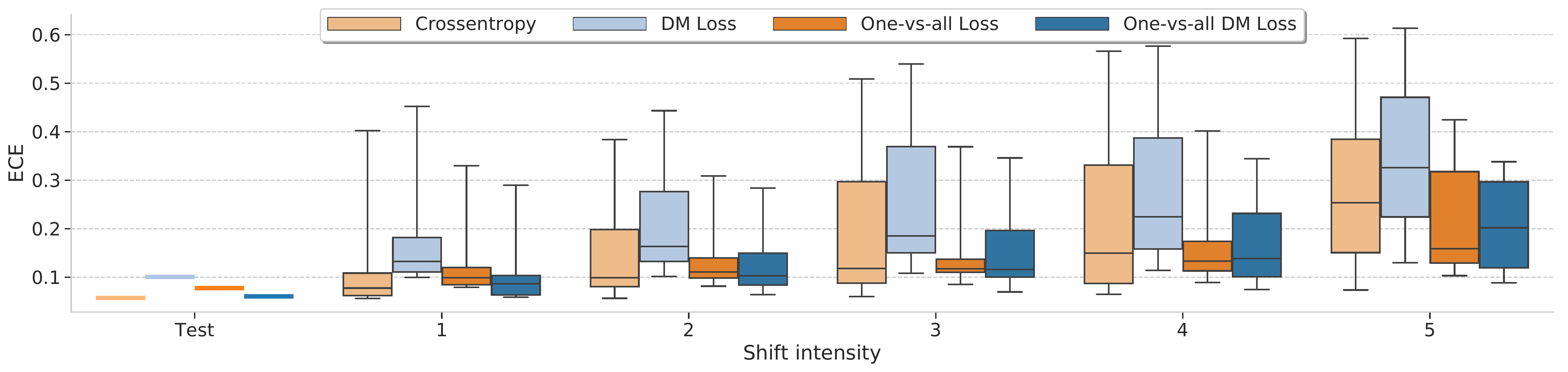}
   } \end{subfigure}
    \begin{subfigure}[Accuracy]{
          \includegraphics[width=0.9\linewidth]{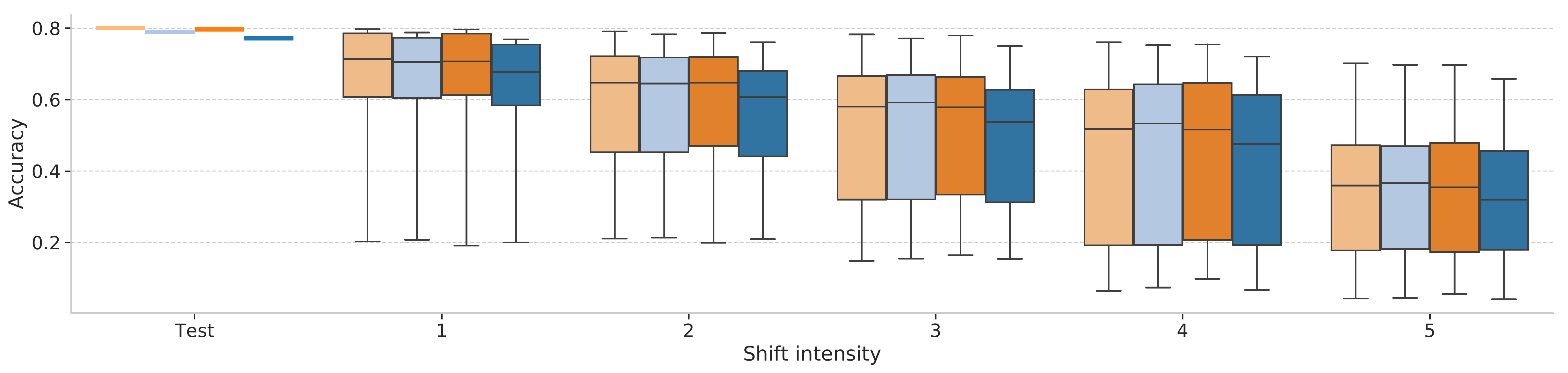}

    }\end{subfigure}
\reducespacebetweenfigureandcaption
    \caption{
    Expected Calibration Error and Accuracy under dataset shift for CIFAR-100-C. One-vs-all classifiers maintain similar predictive accuracy under covariate shift as the softmax cross-entropy baseline, while consistently achieving better calibration under covariate shift, especially as shift increases, and adding a distance-based logit representation further improves ECE performance. Data at \href{https://tensorboard.dev/experiment/YRsq4i4eQDumUqRz41JFkw/\#scalars&_smoothingWeight=0}{this url}.
    }
    \label{fig:accuracy-cifar100c}
\end{figure}

We also report results on the CIFAR-100 dataset using a Wide ResNet 28-10 model \citep{zagoruyko2016wide}, with further implementation details in Appendix \ref{subsection:cifar100}. The results in Figure~\ref{fig:accuracy-cifar100c} show that one-vs-all methods consistently improve uncertainty while maintaining predictive accuracy compared to their softmax-based counterparts, with a distance-based formulation performing the best, especially at high levels of shift.

\subsection{OOD Performance with Accuracy as a Function of Confidence}
\label{subsection:ood_conf}
\begin{figure}[ht]
    \centering
          \includegraphics[width=0.45\linewidth]{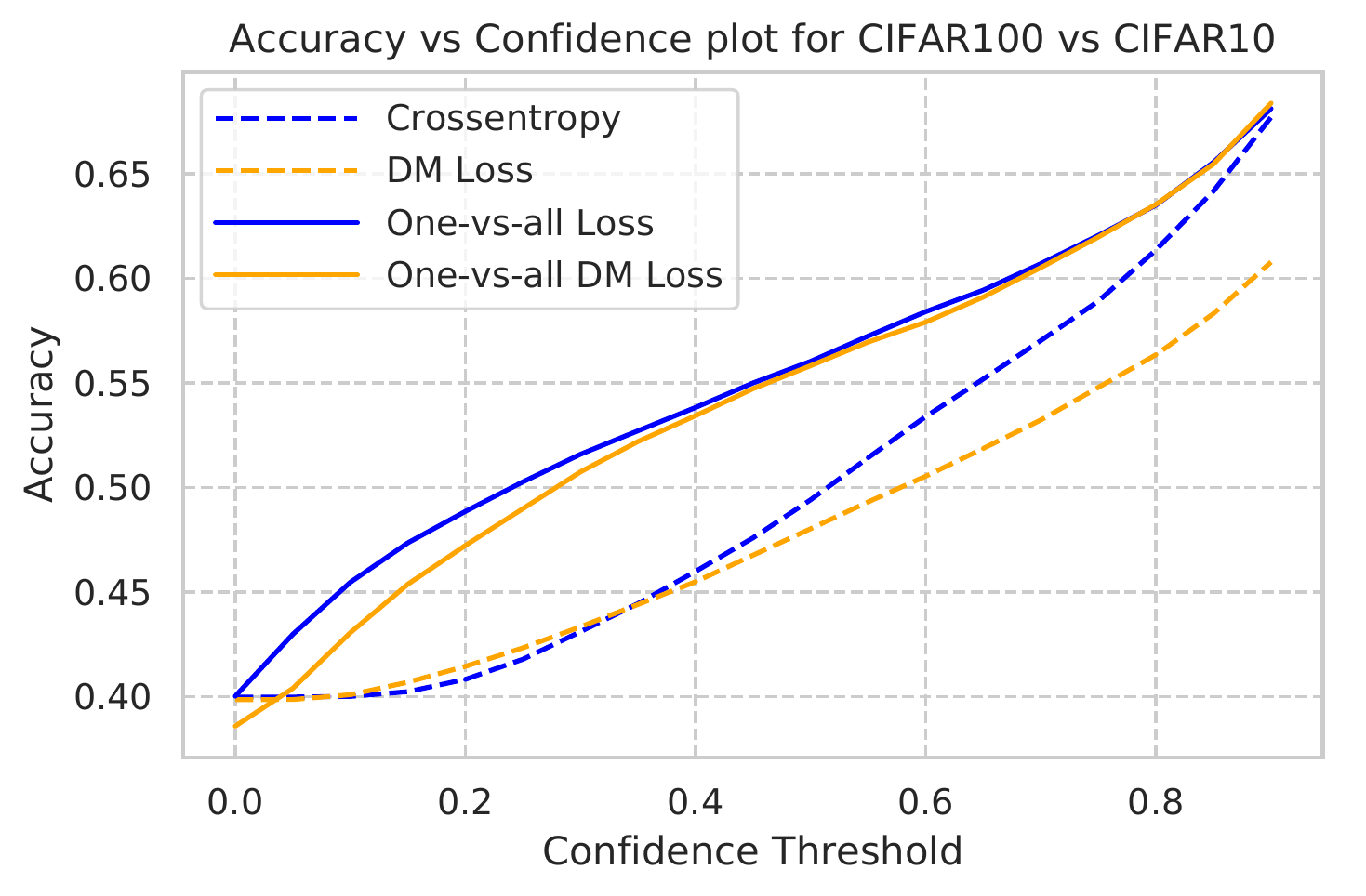}
          \includegraphics[width=0.45\linewidth]{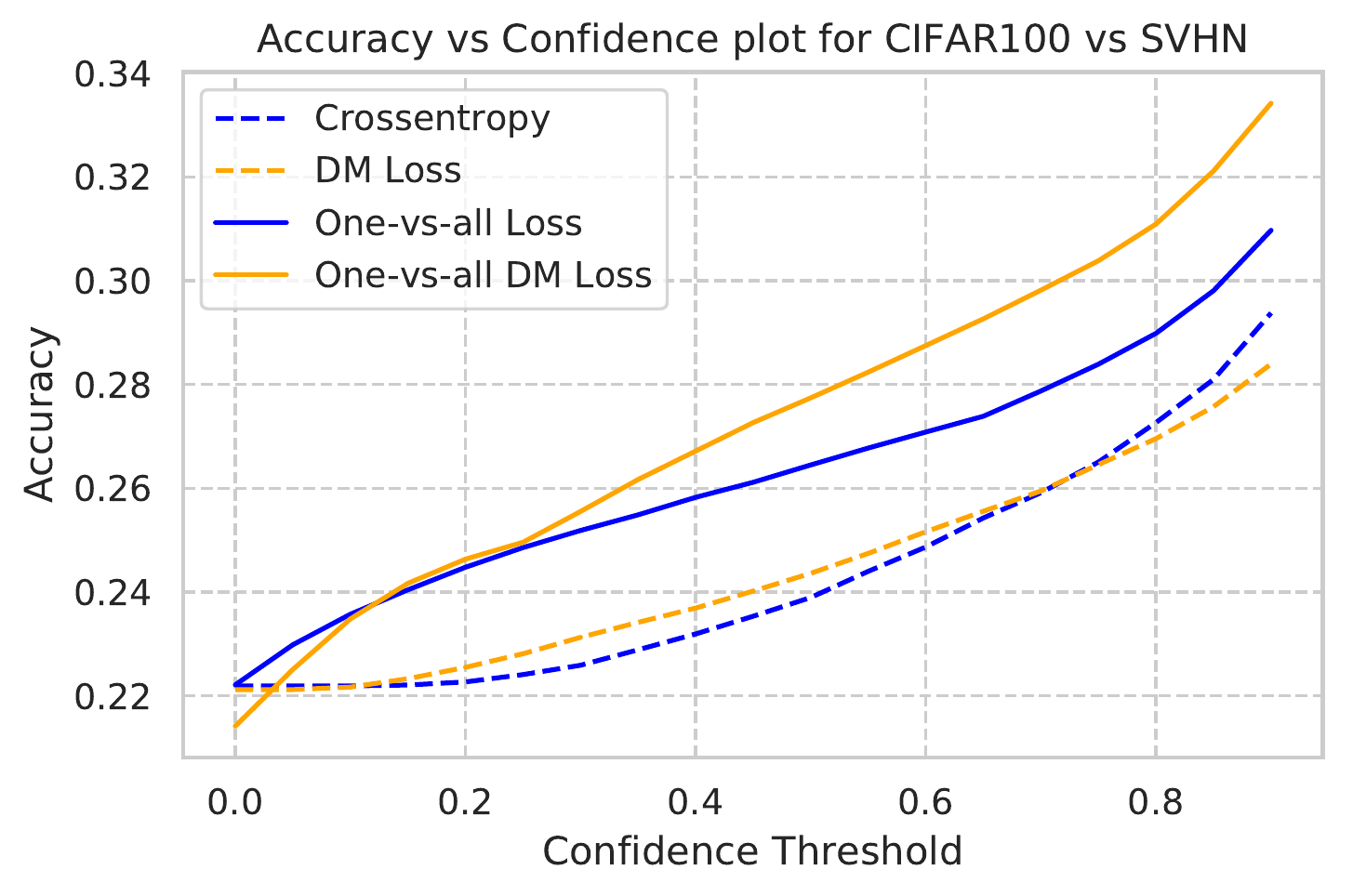}
        \caption{
        {Accuracy vs Confidence curves: Wide ResNet 28-10 model trained on CIFAR-100 and tested on both in-distribution CIFAR-100 test and OOD CIFAR-10 (\textbf{left}) and SVHN (\textbf{right}) test data.} Softmax cross-entropy tends to produce more overconfident incorrect predictions, whereas one-vs-all methods are significantly more robust to OOD data.}
\label{fig:ood-cifar100}
\end{figure}

In order to evaluate the performance of one-vs-all methods in situations where it is preferable for discriminative models to avoid incorrect yet overconfident predictions, we consider plotting accuracy vs. confidence plots \citep{lakshminarayanan2017simple}, where the trained models are evaluated on cases where the confidence estimates are above a user-specified threshold. In Figure~\ref{fig:ood-cifar100}, we plot confidence vs accuracy curves for a Wide ResNet model trained on CIFAR-100, and evaluated on a test set containing both CIFAR-100 and an OOD dataset (SVHN \citep{Netzer2011}, CIFAR-10). Test examples higher than a confidence threshold $0 \leq \tau \leq 1$ are filtered and their accuracy is calculated. From these curves, we would expect a well-calibrated model to have a monotonically increasing curve, with higher accuracies for higher confidence predictions, and we can see that one-vs-all methods consistently result in more robust accuracies across a larger range of confidence thresholds. In Appendix \ref{section:ood_metric}, we also report more traditional OOD detection metrics such as AUROC and AUPRC, and show how one-vs-all DM loss performs poorly due to a high overlap between incorrectly classified in-distribution points and OOD points, especially at lower confidences.

\section{Conclusions}
In order to improve the predictive uncertainty of discriminative models, we analyze the different failure modes of softmax crossentropy, and show that the softmax function contributes significantly to overconfident predictions under covariate shift and on OOD data. We propose a one-vs-all formulation instead that improves the performance of neural networks on image classification tasks. We also show how the linear transformations of the embeddings that result in the logits also contribute to miscalibration, and demonstrate how distance-based logits followed by a softmax normalization do not mitigate this issue. However, combining a distance-based formulation with a one-vs-all scheme results in even better performance on certain image datasets, and on a language intent classification task (Appendix~\ref{subsection:intent_results}).
However, one-vs-all distance-based losses suffer from a few setbacks; we observe that training from scratch is challenging for datasets with a large value of $K$ such as Imagenet (Appendix~\ref{subsection:imagenet_results}), and further exploration is required to scale these losses to such large datasets, and to non-image domains. However, we believe that decomposing a multiclass problem into binary tasks using one-vs-all formulations allows for further exploration into binary losses that have been extensively studied for calibration tasks \citep{gneiting2007strictly, reid2010composite, mukhoti2020the}, which would make for an interesting avenue of further research.

\clearpage\newpage
\bibliography{references.bbl}
\bibliographystyle{icml2020}

\appendix
 \clearpage\newpage
\appendix
 \setcounter{figure}{0}
\setcounter{table}{0}
\makeatletter 
\renewcommand{\thefigure}{S\@arabic\c@figure}
\renewcommand{\thetable}{S\@arabic\c@table}
\makeatother

\section{Comparison of learned class centers}
\label{section:centers}
\begin{figure}[ht]
\centering
          \includegraphics[width=0.48\linewidth]{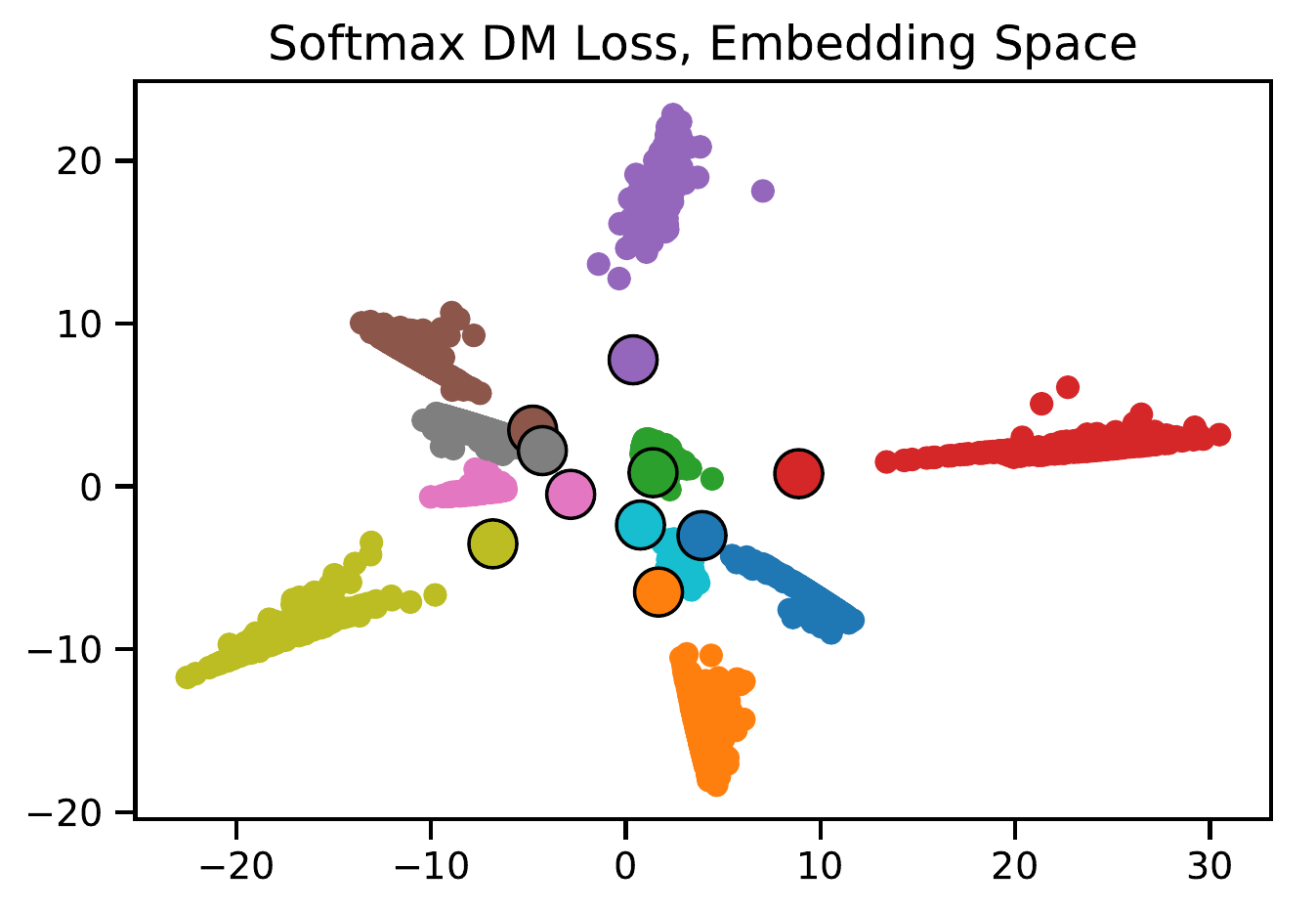}
          \includegraphics[width=0.48\linewidth]{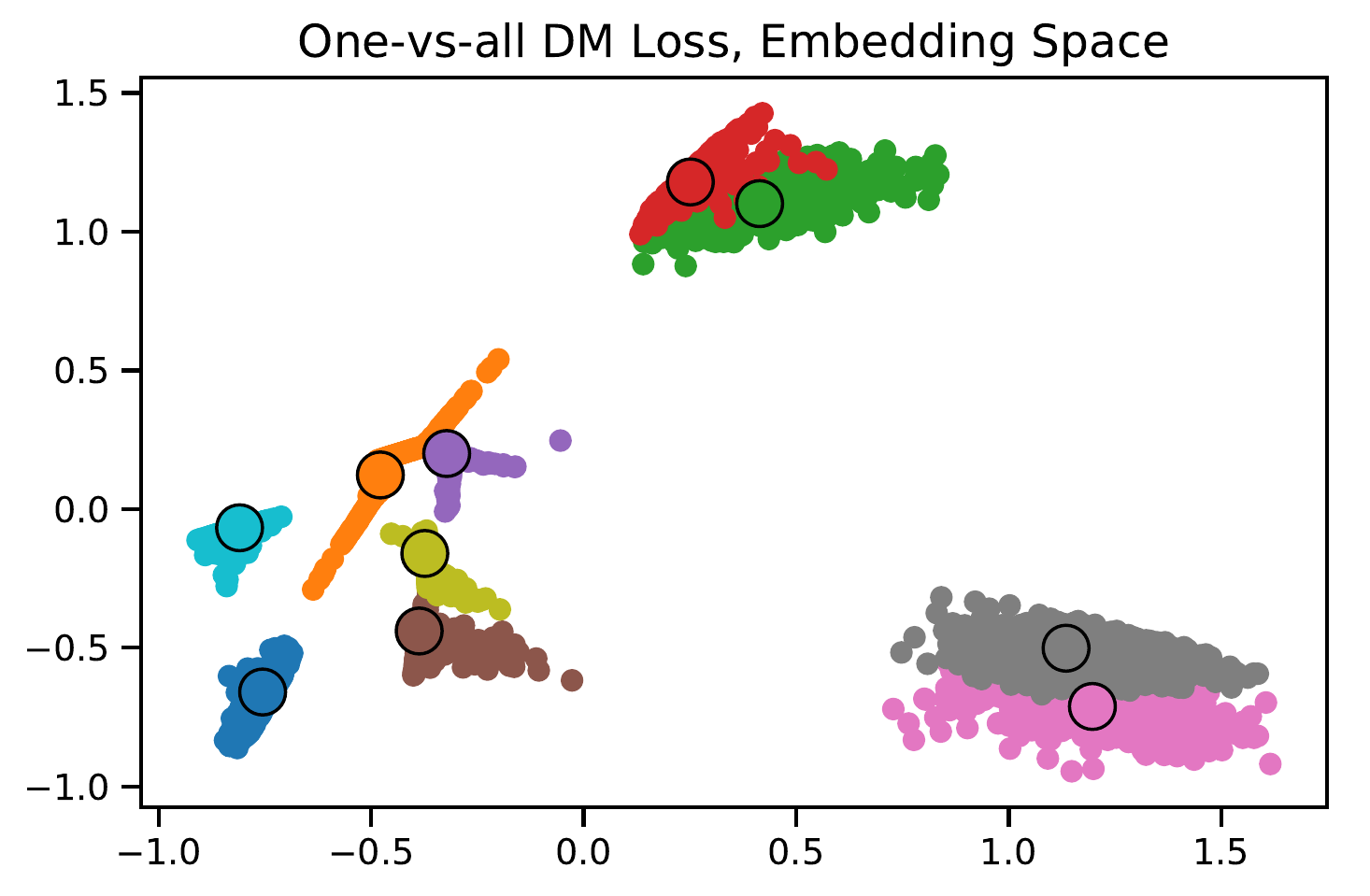}
        \reducespacebetweenfigureandcaption \caption{
       Penultimate Layer Embeddings and Learned Class Centers (solid circles) for DM Loss \textbf{(left)} and One-vs-all DM Loss \textbf{(right)}. It can be seen that the weights of the last layer learnt by DM Loss do not accurately represent the class centers of the embeddings, due to the softmax normalization. However, the one-vs-all formulation learns more meaningful class-centers that better approximate the centers of class-specific embeddings. Implementation details are mentioned in Appendix~\ref{subsection:2d}. 
        }
        \label{fig:classcenters}
\end{figure}

In order to better portray the optimization issues introduced by the softmax normalization in learning meaningful class centers using a DM Loss formulation, we train models on a 2D toy dataset using distance-based logits, but with differing normalization schemes. In both cases, the learnt embedding space is 16-dimensional, and the weight matrix of the last layer is given by $\bm{W} \in \mathbb{R}^{16 \times 10}$, where the $j$th column $\bm{W}_{:,j}$ corresponds to the learned class center for the $j$th class. In order to plot the learned class centers in Figure~\ref{fig:classcenters}, the 16-dimensional hidden embedding space obtained from a forward pass of the training data is projected onto a 2-dimensional PCA subspace, along with the columns of the weight matrix of the last layer.

\section{Reliability Diagrams}
\label{section:reliability}
\begin{figure}[ht]
\centering
          \includegraphics[width=0.48\linewidth]{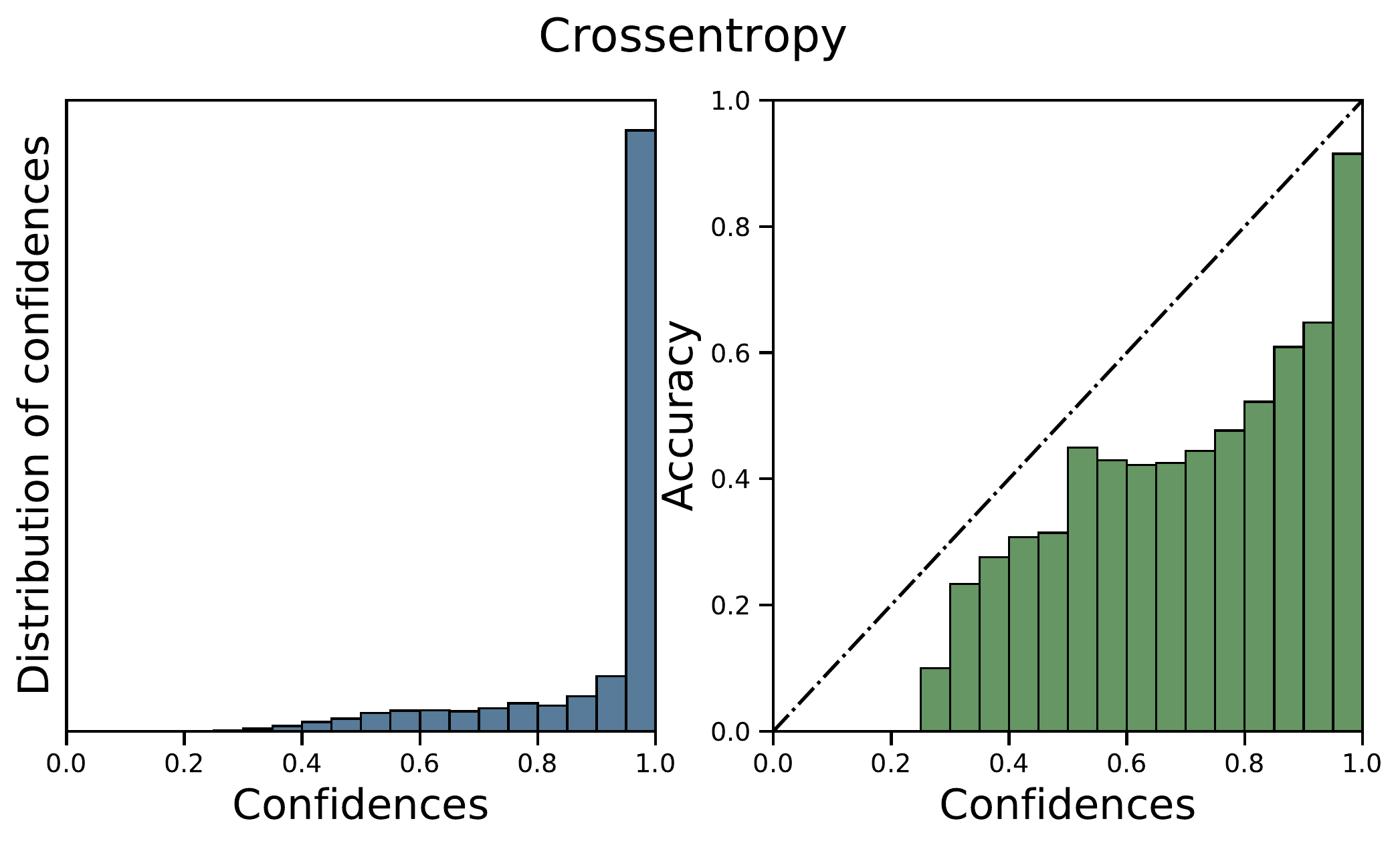}
          \includegraphics[width=0.48\linewidth]{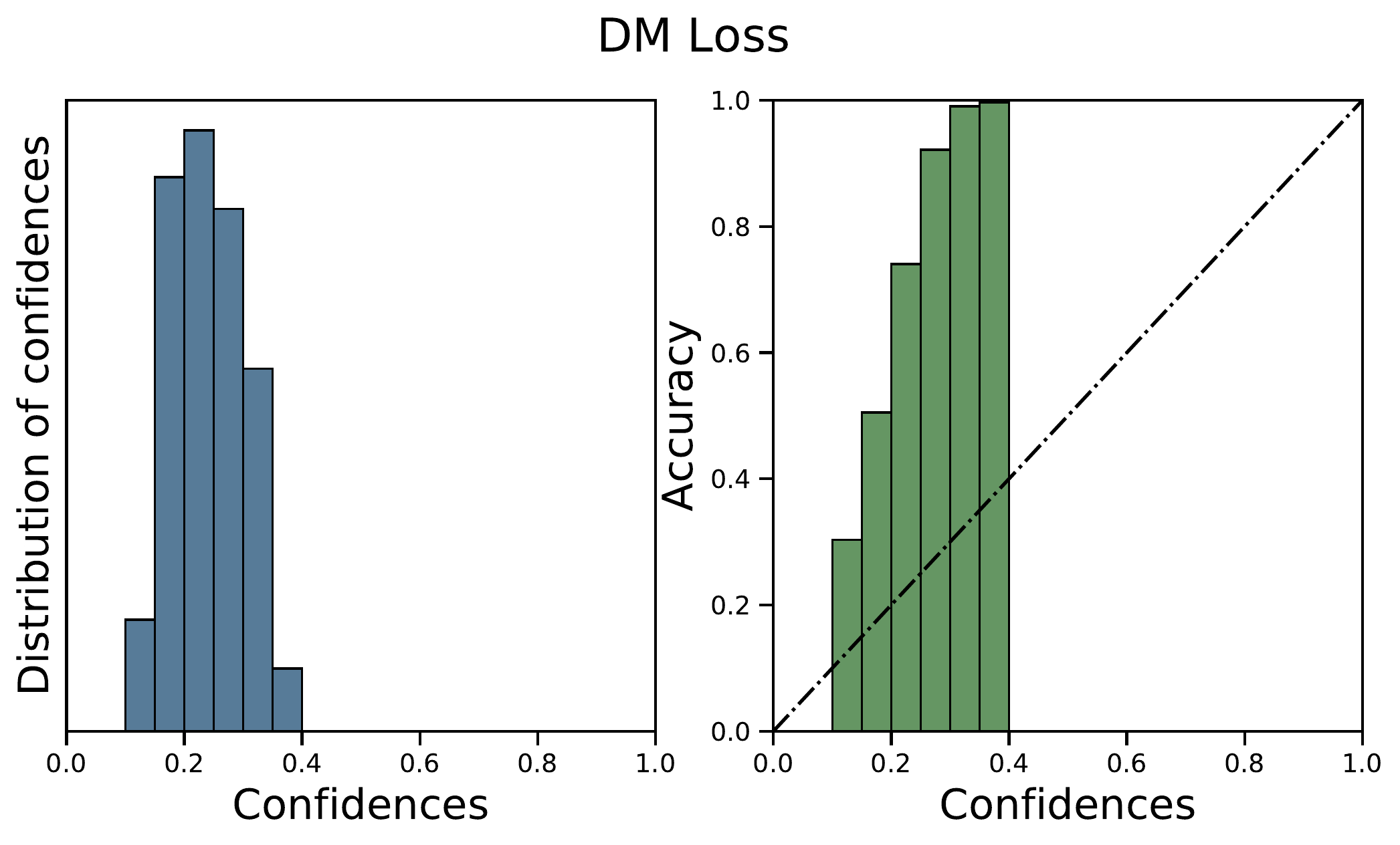}
          \includegraphics[width=0.48\linewidth]{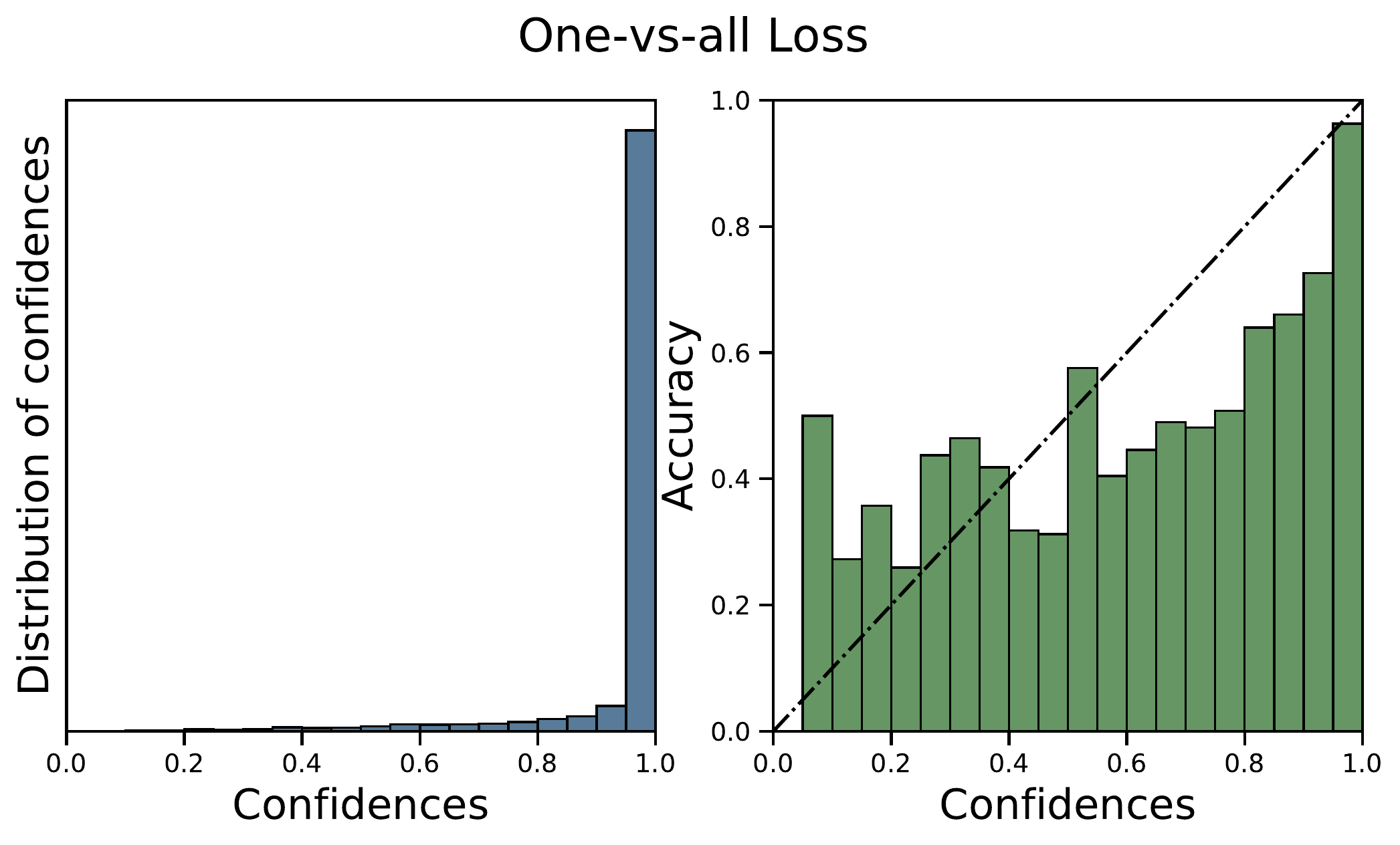}
          \includegraphics[width=0.48\linewidth]{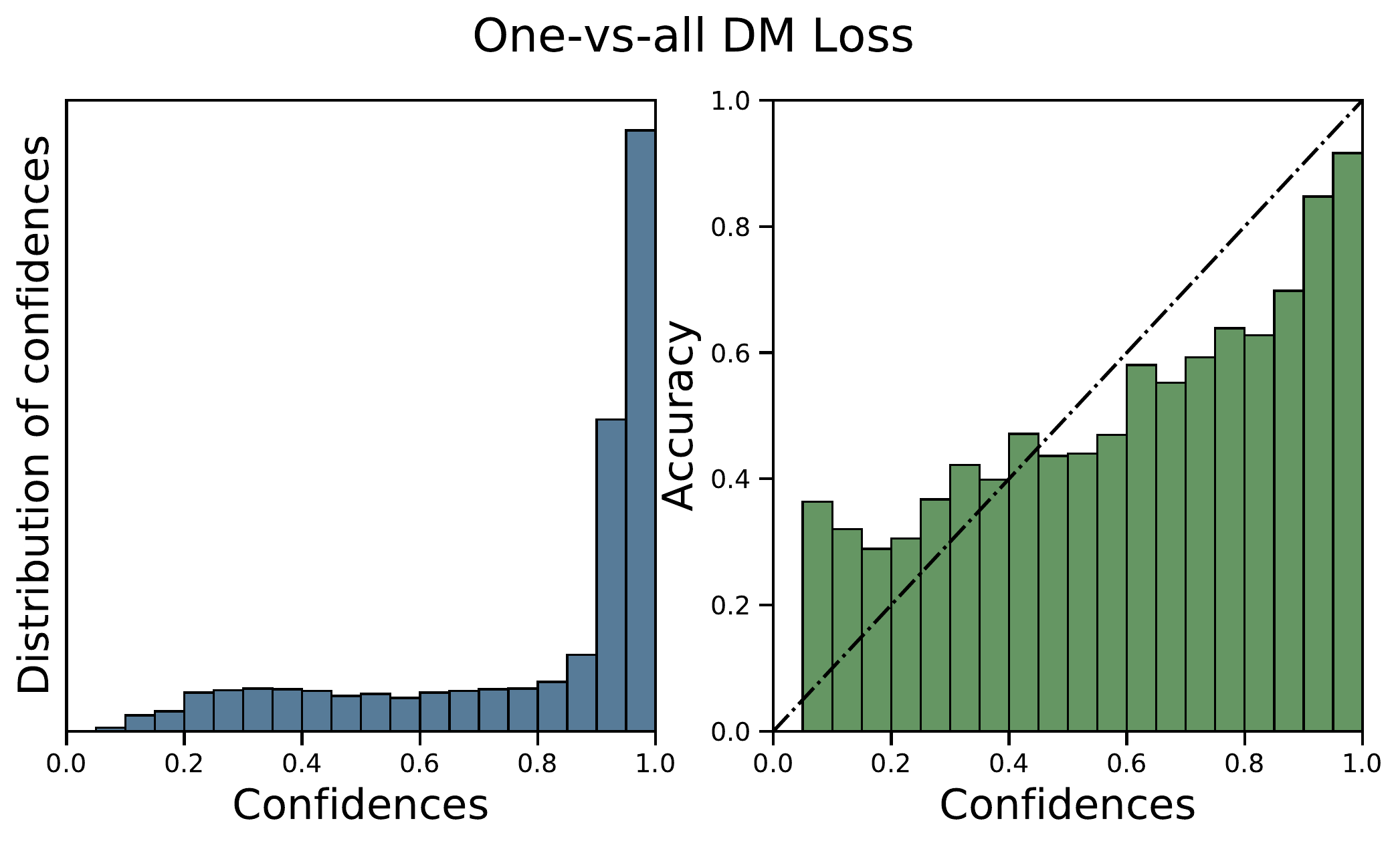}
        \reducespacebetweenfigureandcaption \caption{
       Reliability diagrams for ResNet 20 models trained on the CIFAR-10 dataset, for softmax cross-entropy \textbf{(top-left)}, DM loss \textbf{(top-right)}, One-vs-all loss \textbf{(bottom-left)} and One-vs-all DM loss \textbf{(bottom-right)}. One-vs-all methods show better-calibrated reliability curves, especially showing a larger number of points with lower confidence on the in-distribution test set, with correspondingly low accuracies. Of particular note is the behavior of DM loss, where there is a systematic issue of underconfidence, as all points lie below a threshold. This behavior has been reported in \citet{macedoisotropic}, and though this results in better OOD performance overall, the exceedingly low confidences result in very miscalibrated predictions in-distribution and under dataset shift.
        }
        \label{fig:rel_cifar10}
\end{figure}

\begin{figure}[ht]
\centering
          \includegraphics[width=0.48\linewidth]{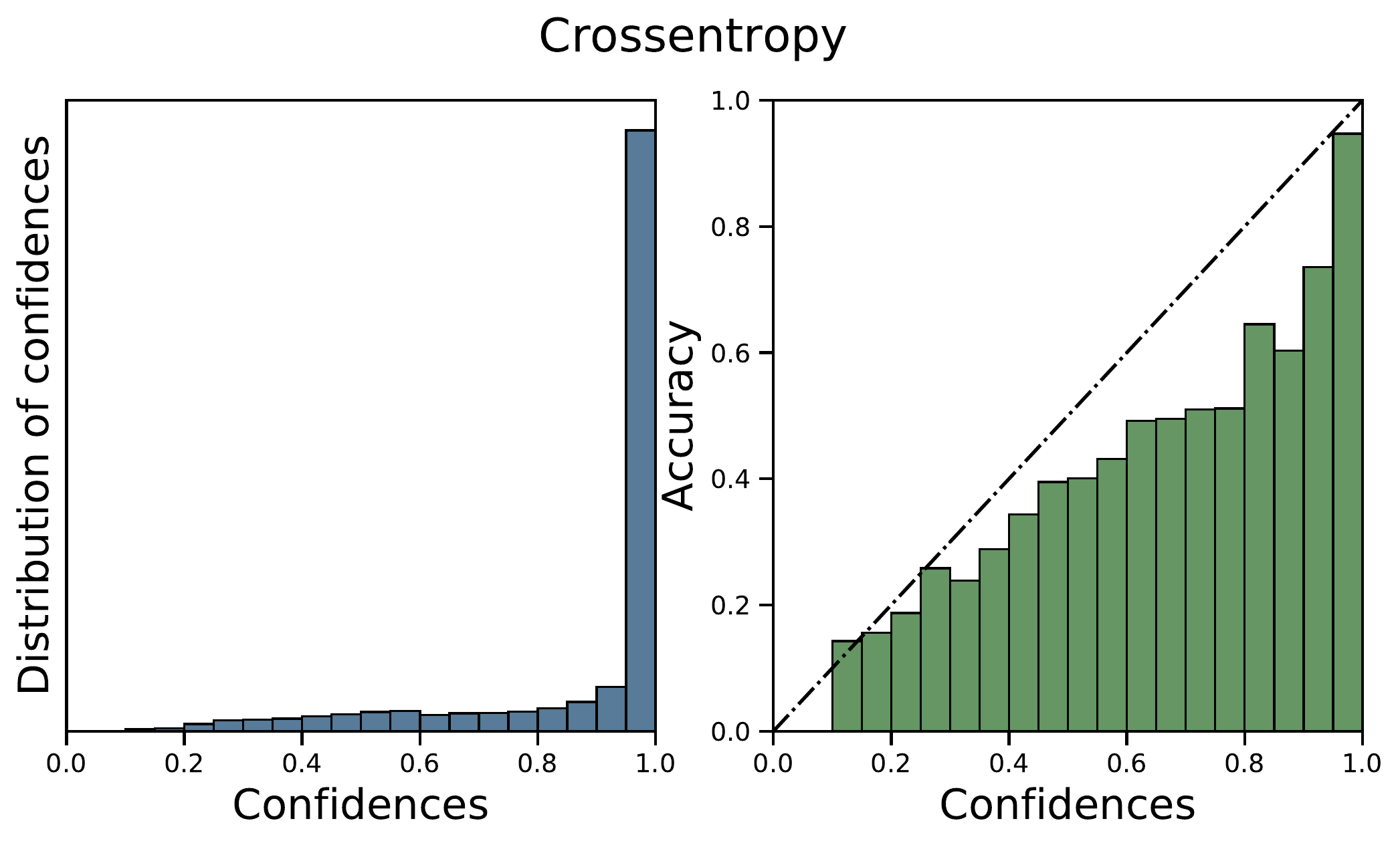}
          \includegraphics[width=0.48\linewidth]{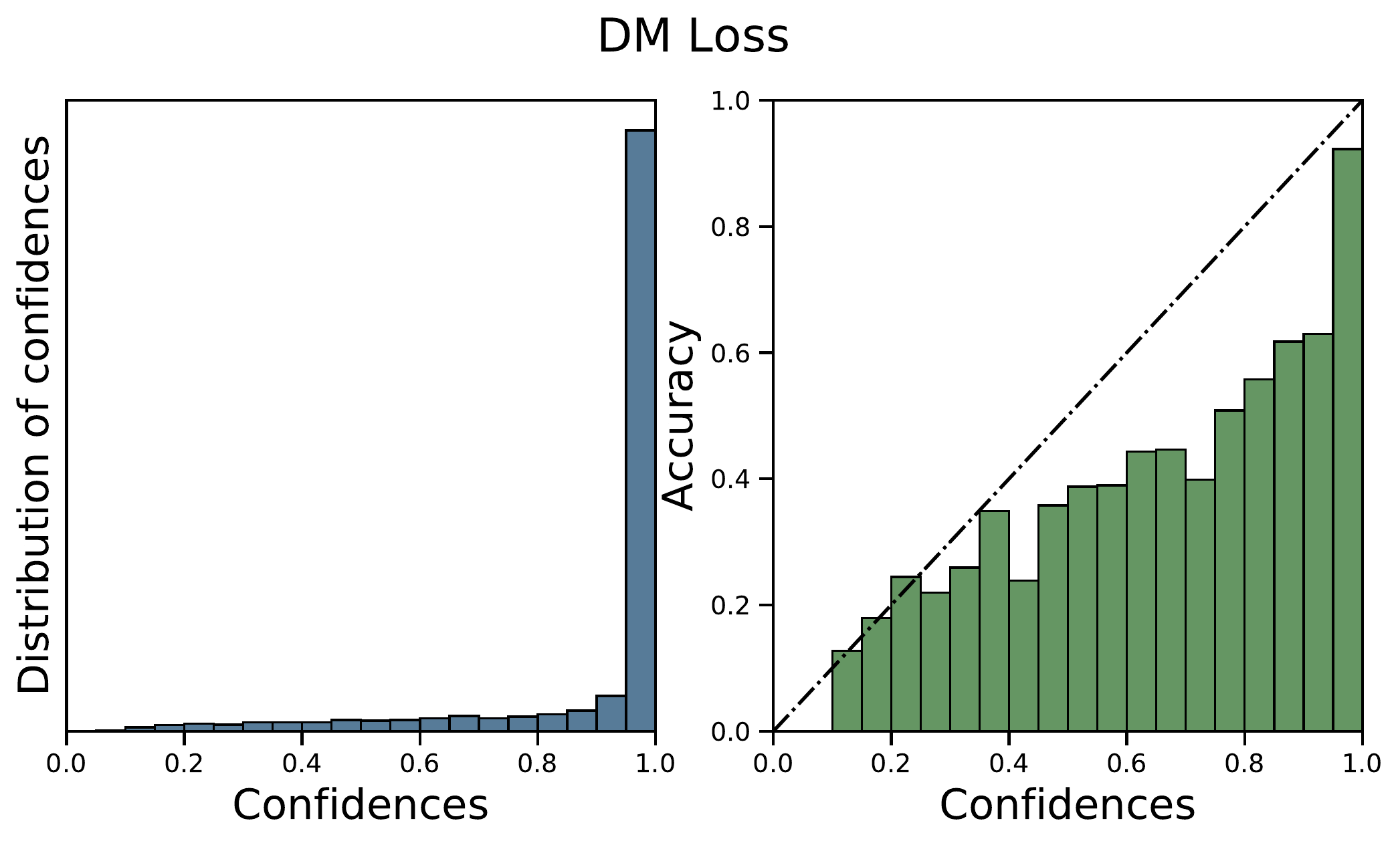}
          \includegraphics[width=0.48\linewidth]{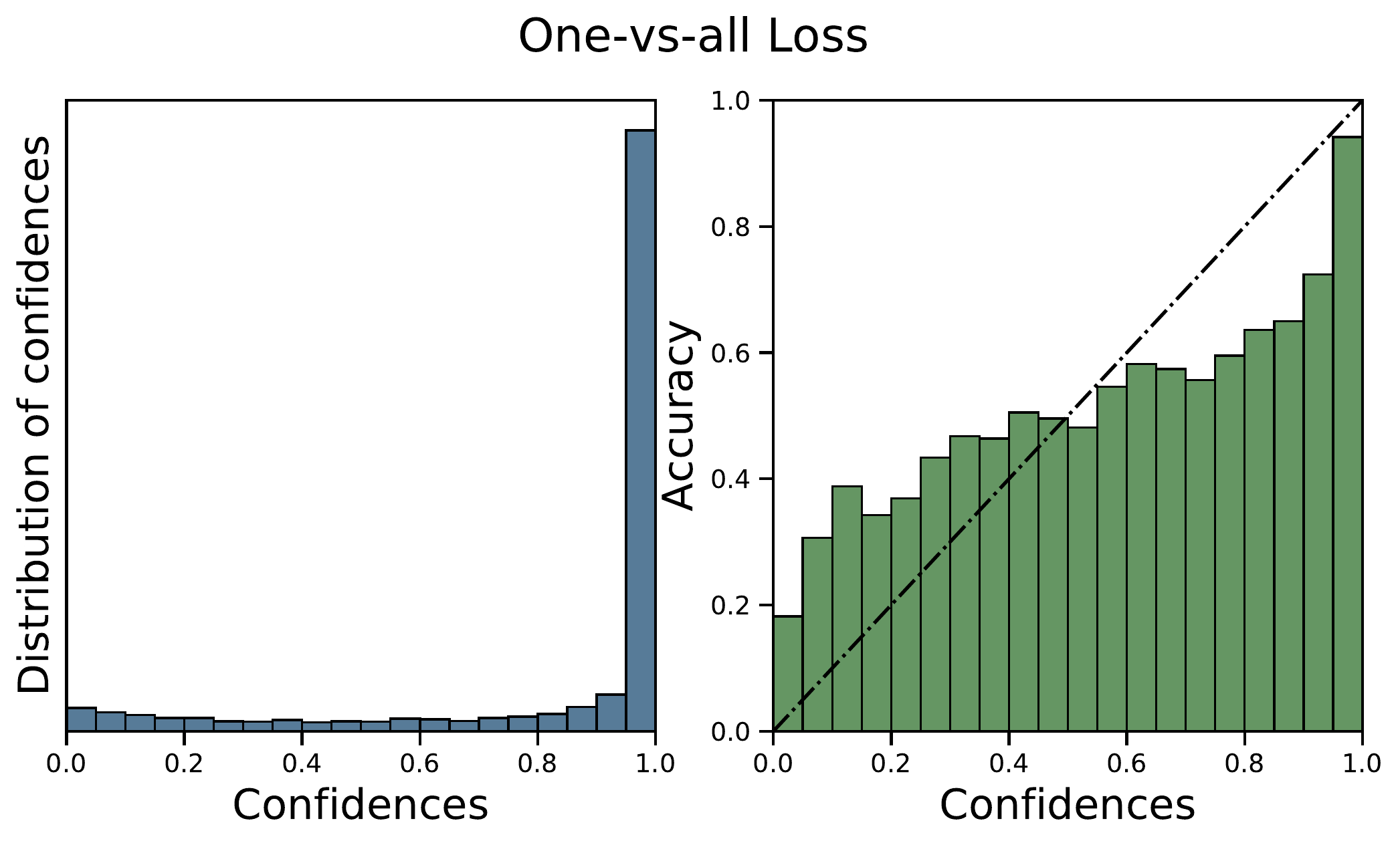}
          \includegraphics[width=0.48\linewidth]{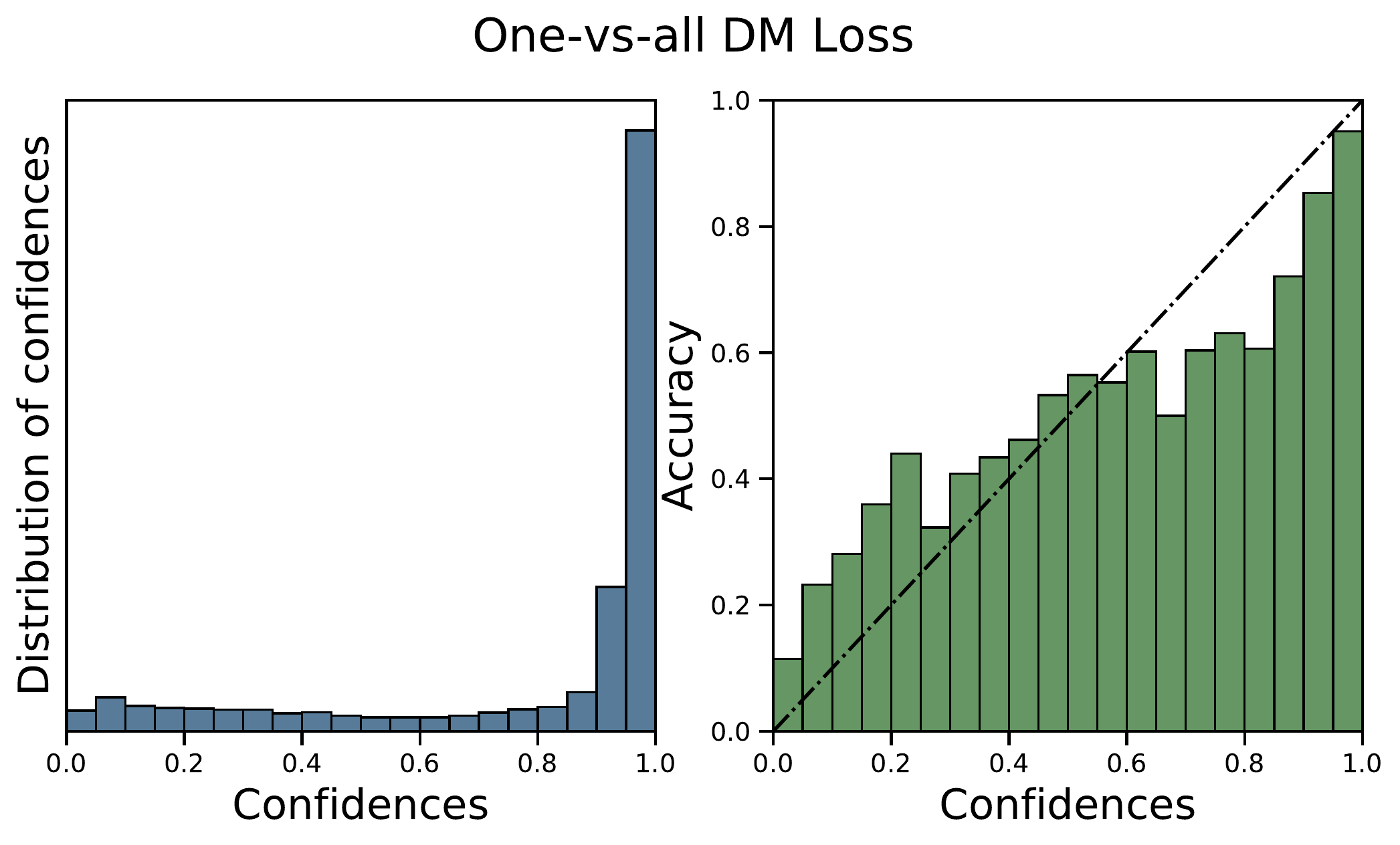}
        \reducespacebetweenfigureandcaption \caption{
       Reliability diagrams for Wide ResNet 28-10 models trained on the CIFAR-100 dataset, for softmax cross-entropy \textbf{(top-left)}, DM loss \textbf{(top-right)}, One-vs-all loss \textbf{(bottom-left)} and One-vs-all DM loss \textbf{(bottom-right)}. One-vs-all methods consistently have more datapoints with lower confidences, which correspond to lower accuracy predictions as well, as opposed to the softmax-based methods. Interestingly, DM loss does not suffer the same issue of underconfidence as in CIFAR-10 (Figure~\ref{fig:rel_cifar10}).
        }
        \label{fig:rel_cifar100}
\end{figure}

To further understand the interplay of behaviors between one-vs-all normalizations and distance-based logits, we plot reliability diagrams \citep{degroot1983comparison, niculescu2005predicting} which show accuracy as a function of confidence in the form of binned histograms of confidences over the in-distribution test set. We plot reliability diagrams for models trained on CIFAR-10 in Figure~\ref{fig:rel_cifar10} and CIFAR-100 in Figure~\ref{fig:rel_cifar100}. From these plots, we can see that softmax cross-entropy tends to result in models that have very high confidences on the test set, often with lower accuracies than expected. One-vs-all methods tend to give better calibrated predictions, especially for points with lower accuracy corresponding to lower confidence predictions. Interestingly, we see somewhat anomalous behavior of DM loss on the CIFAR-10 dataset, which has been shown before in \citet{macedoisotropic}, where all confidences for points in the in-distribution test set are lower than a certain threshold. One hypothesis for this behavior is the fact that it is easy to obtain high accuracy on the training data without necessarily learning to cluster points close to their class centers, as long as all points of a class are closest to the corresponding weight in the last layer, and this sort of behavior is observed even in the 2D toy example in Appendix~\ref{section:centers}. This anomalous histogram of confidences also explains the decreasing trend of ECE seen under dataset shift in Figure~\ref{fig:accuracy-cifar10c}, as under higher intensity of shift, models would tend to get less accurate, while the predictions made by a model trained with the DM loss are consistently underconfident. We do not see this behavior for CIFAR-100 however.
\section{Additional Results}
\subsection{ImageNet}
\label{subsection:imagenet_results}
We report the performance of the proposed loss functions on the ImageNet dataset \citep{russakovsky2015imagenet}, using a ResNet-50 model \citep{he2016deep} trained using the setup described in Appendix \ref{subsection:imagenet}. From Figure~\ref{fig:acc-imagenetc}, we see that a one-vs-all formulation with affine-transformed logits has similar predictive accuracy to softmax cross-entropy, and performs more robustly under dataset shift, especially at higher intensities of shift. The distance-based one-vs-all formulation however has difficulty matching the predictive performance of softmax cross-entropy, while still having lower ECE under dataset shift. We hypothesize that the large number of classes in ImageNet make a distance-based formulation difficult to scale, especially since the gradients to learn class centers are calculated in a mini-batch setting, and even with a batch size of 4096, the number of points per class can be heavily imbalanced, resulting in poor gradients for optimization.

\begin{figure}[ht]
    \centering
          \includegraphics[width=0.48\linewidth]{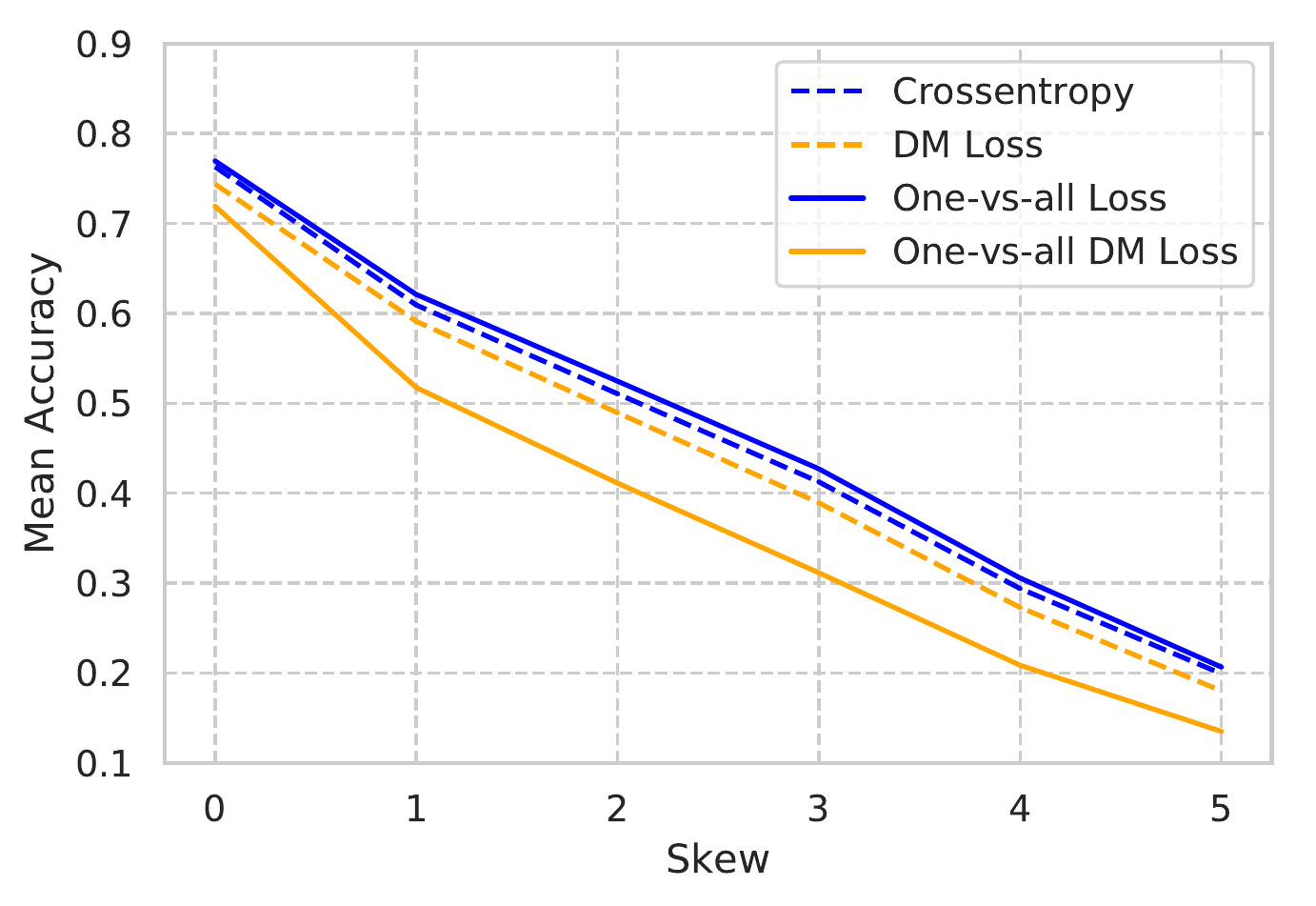}
          \includegraphics[width=0.48\linewidth]{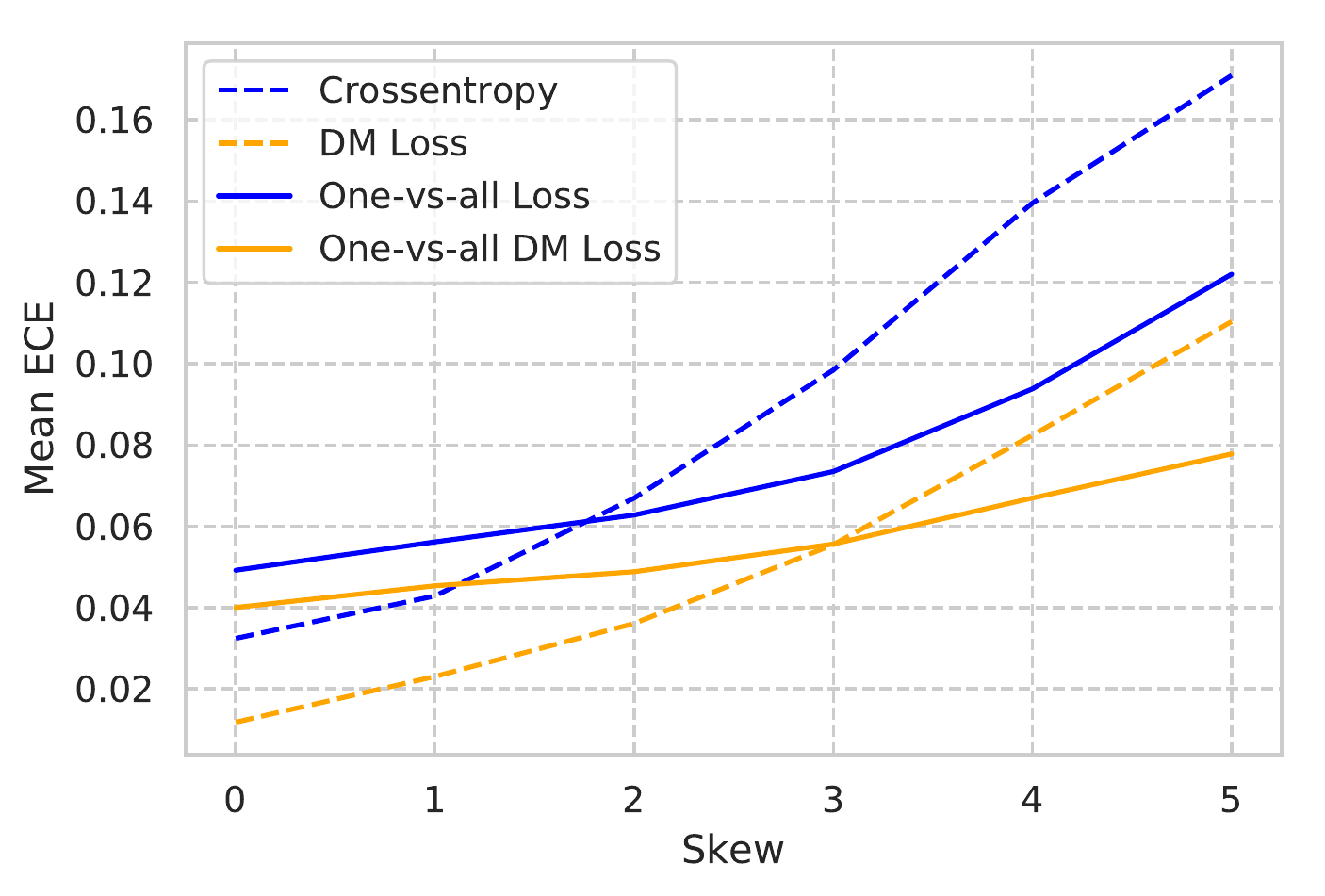}
         \reducespacebetweenfigureandcaption 
        \caption{%\textbf{
        Mean Accuracy (left) and Mean Expected Calibration Error (right) for a ResNet-50 model trained on  ImageNet and evaluated on the ImageNet-C corrupted data, under 5 levels of shift.
        }
\label{fig:acc-imagenetc}
\end{figure}

\subsection{CLINC Intent Classification Dataset}
\label{subsection:intent_results}
\begin{figure}[ht]
    \centering
          \includegraphics[width=0.75\linewidth]{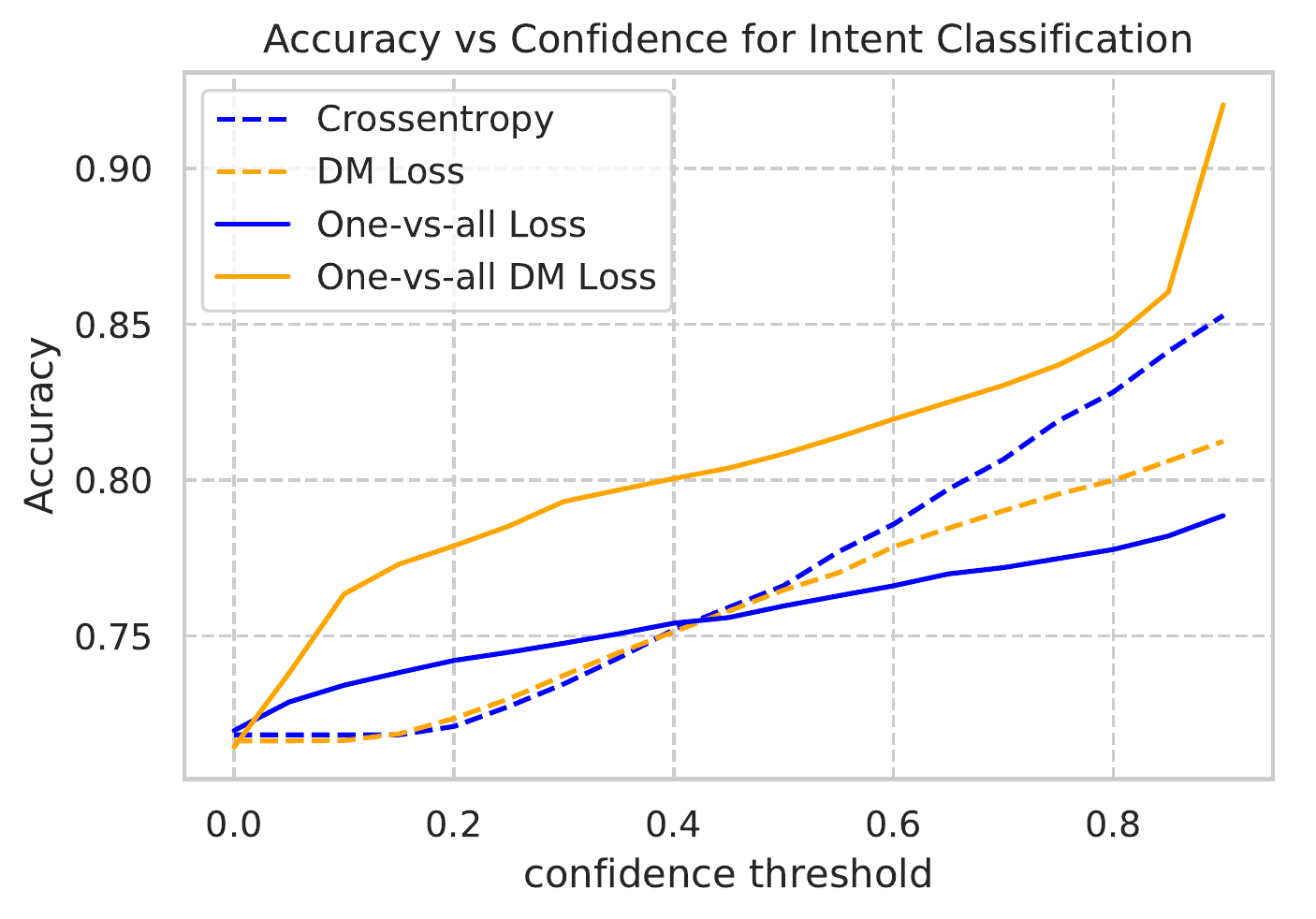}
        \caption{%\textbf
        Accuracy vs Confidence curves: \texttt{BERT-Mini} model trained on the CLINC intent classification dataset. Softmax crossentropy tends to produce more overconfident incorrect predictions, whereas one-vs-all DM loss is significantly more robust to out-of-distribution data, especially at lower confidence thresholds. }
\label{fig:ood-bert}
\end{figure}

We also evaluate the performance of one-vs-all methods on out-of-distribution data on modalities beyond common image tasks, namely a language understanding task using the CLINC out-of-scope intent detection benchmark dataset \citep{larson2019evaluation}. The benchmark dataset contains 150 types of in-domain services and 1500 out-of-domain sentences, and it is important for real-world, goal-oriented dialog systems to be able to distinguish between in-domain and out-of-domain sentences. For this task, we train a \texttt{BERT-Mini} model \citep{devlin2018bert, tsai2019small} on the in-domain data from scratch (with further details in Appendix \ref{subsection:intent}), and report both the in-distribution ECE, and performance on out-of-distribution data. From Table~\ref{tab:clinc}, we see that one-vs-all formulations are competitive with the predictive accuracy achieved by softmax cross-entropy, while improving in-distribution calibration. Simultaneously, we can see from Figure~\ref{fig:ood-bert} that one-vs-all DM Loss performs more robustly at different confidence thresholds, however regular one-vs-all loss struggles at higher confidences to distinguish in-distribution sentences from OOD sentences.
\begin{table}[h]
\centering
\begin{tabular}{l|l|l}
\textbf{Loss function} & \textbf{Test Accuracy} & \textbf{Test ECE} \\ \hline
Crossentropy           & 90.16                & 0.079             \\
DM Loss             & 89.89                  & 0.083            \\
One-vs-all Loss            & \textbf{90.67}                  & 0.078            \\
One-vs-all DM Loss        & 88.89                  & \textbf{0.052}    
\end{tabular}
\label{tab:clinc}
\caption{In-distribution test accuracy and test ECE for a \texttt{BERT-Mini} model trained on the Intent Classification dataset. Results are averaged over 5 runs.}
\end{table}

\section{On OOD detection metrics: AUROC versus Confidence-Accuracy curves}
\label{section:ood_metric}
In this section, we report the Area Under the Receiver Operating Characteristic curve (AUROC) and Area under the Precision-Recall curve (AUPRC) for OOD tasks in the image domain. Both the AUROC and AUPRC are calculated for the binary task of classifying in-distribution test points from OOD test points. From Tables~\ref{tab:ood_cifar10} and  \ref{tab:ood_cifar100}, we can see that one-vs-all DM loss performs poorly compared to the other methods for a scalar metric such as AUROC and AUPRC. However, from the confidence versus accuracy plots in Section \ref{subsection:ood_conf}, we can see that one-vs-all DM loss is more sensitive to OOD inputs, especially at lower confidence values.
In order to explain this apparent disparity, we  plot the histograms of the predicted confidence for three sets of points: 
\begin{enumerate}[itemsep=0ex]
    \item correctly classified in-distribution test points, 
    \item incorrectly classified in-distribution test points, and 
    \item OOD test points.
\end{enumerate}  

\begin{figure*}[ht]
    \centering
    \begin{subfigure}[CIFAR-10 vs CIFAR-100]{
          \includegraphics[width=0.475\linewidth]{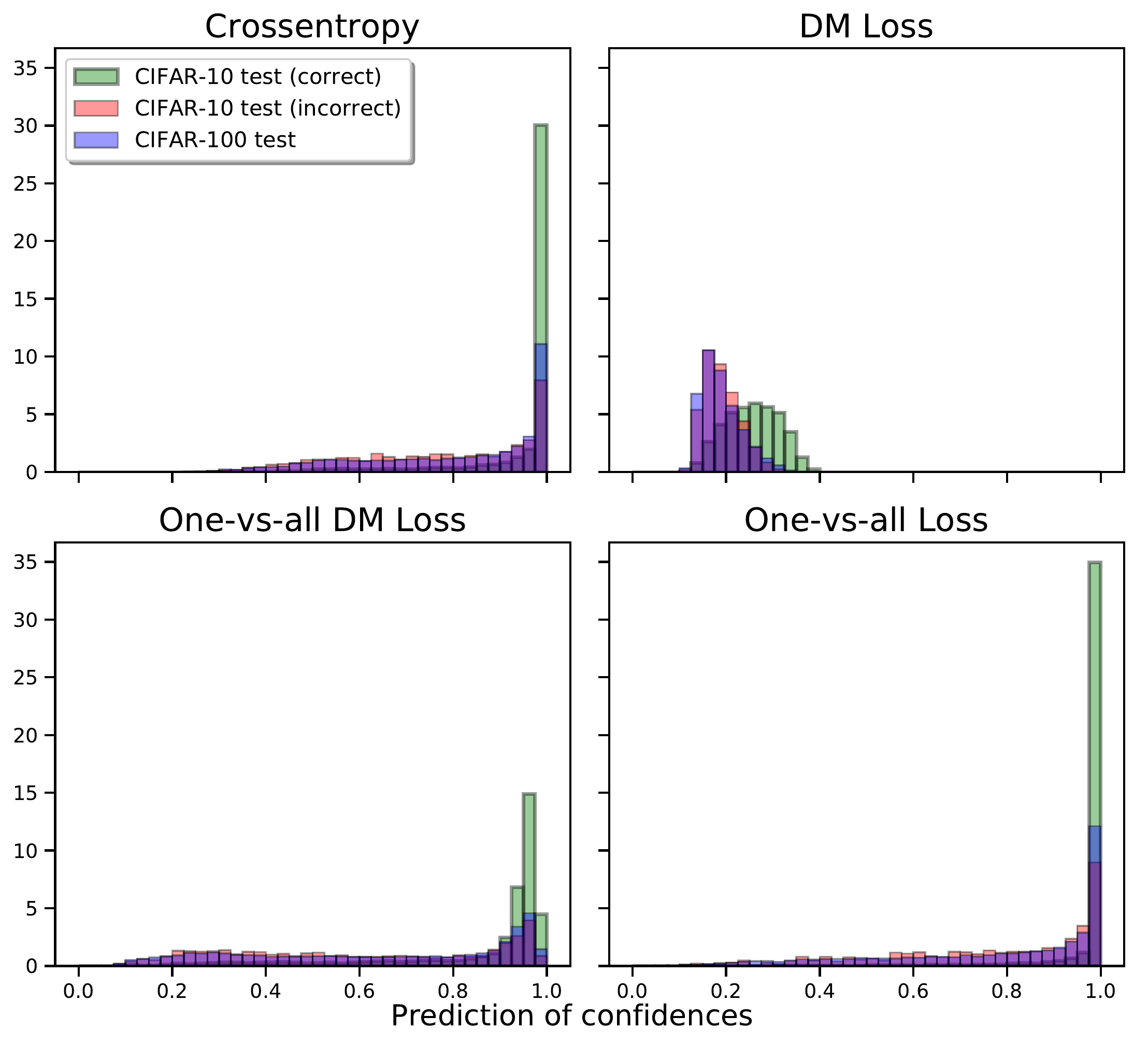}
         \reducespacebetweenfigureandcaption 
    } \end{subfigure}
    \begin{subfigure}[CIFAR-10 vs SVHN]{
          \includegraphics[width=0.475\linewidth]{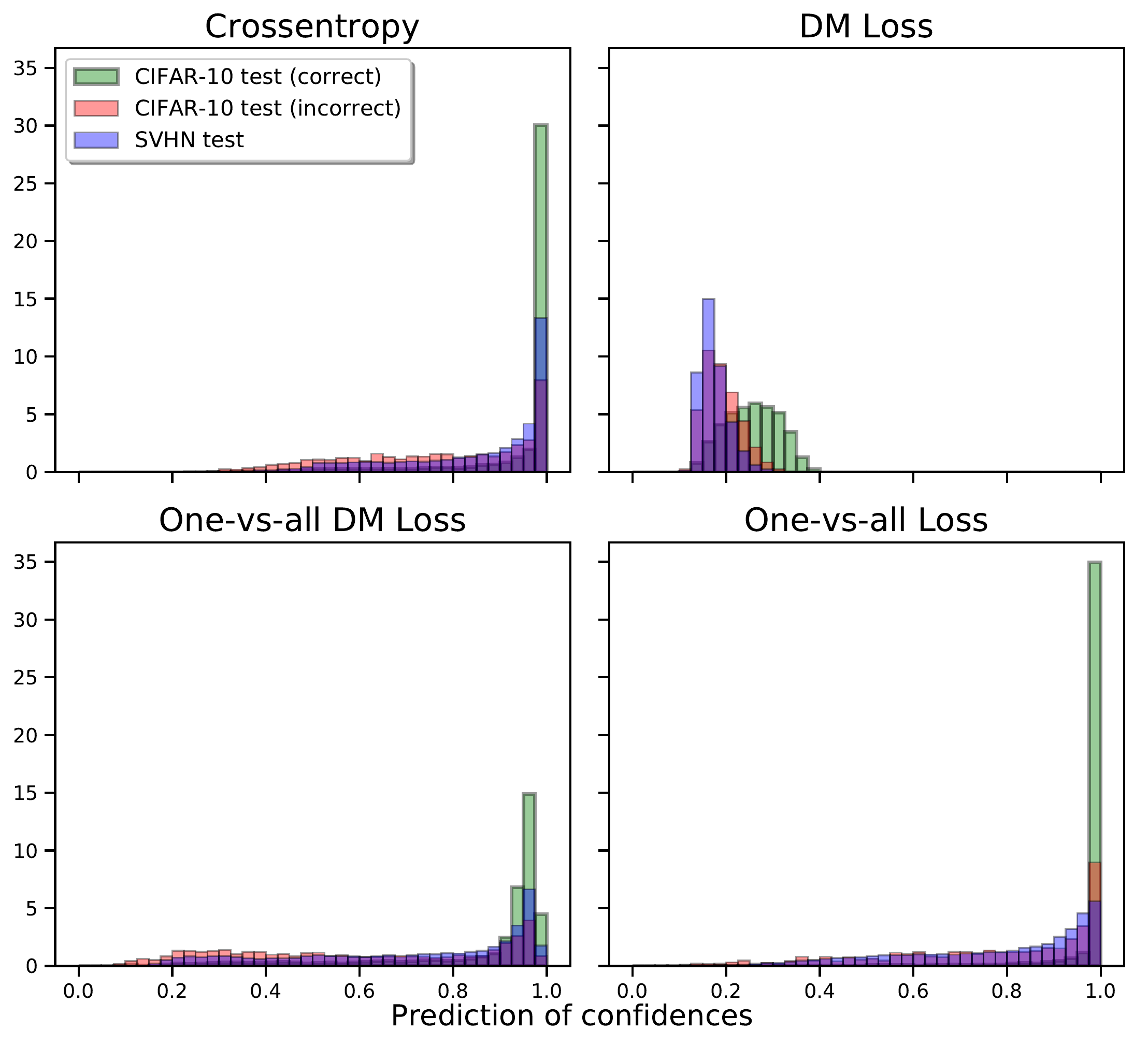}
         \reducespacebetweenfigureandcaption 
    } \end{subfigure}
\caption{%\textbf{
        Histograms of predicted confidences for a ResNet-20 architecture trained with different loss functions on the CIFAR-10 dataset, and tested on the OOD CIFAR-100 and SVHN datasets. We can see that there is a much higher overlap in the confidence distributions of incorrectly classified in-distribution test points and OOD test points, which results in poorer performance when using a scalar metric such as AUROC and AUPRC that aggregate over all thresholds in $[0, 1]$.}
        \label{fig:ood_conf_cifar10_svhn}
\end{figure*}

\begin{figure*}[ht]
    \centering
    \begin{subfigure}[CIFAR-100 vs CIFAR-10]{
          \includegraphics[width=0.475\linewidth]{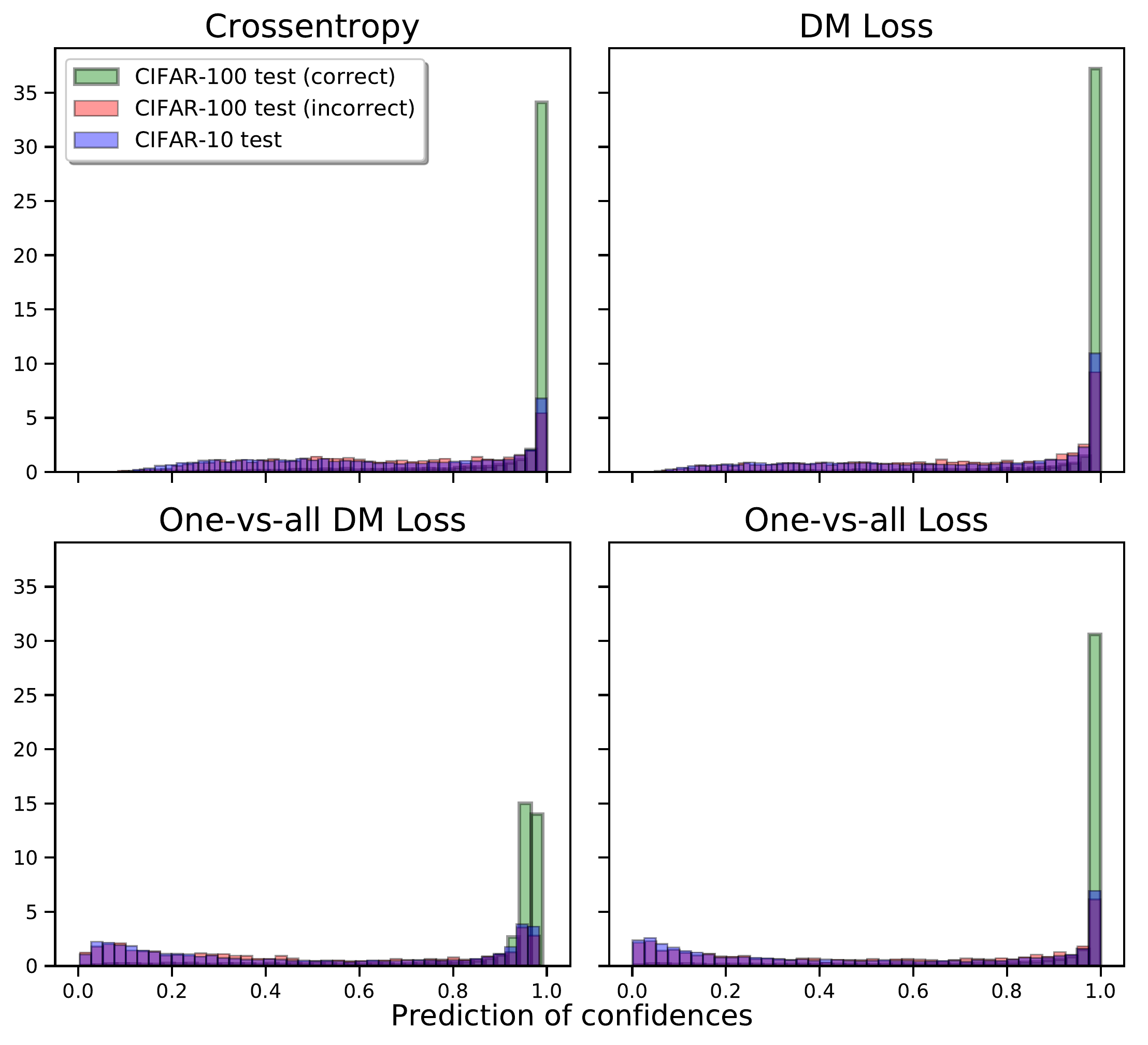}
         \reducespacebetweenfigureandcaption 
    } \end{subfigure}
    \begin{subfigure}[CIFAR-100 vs SVHN]{
          \includegraphics[width=0.475\linewidth]{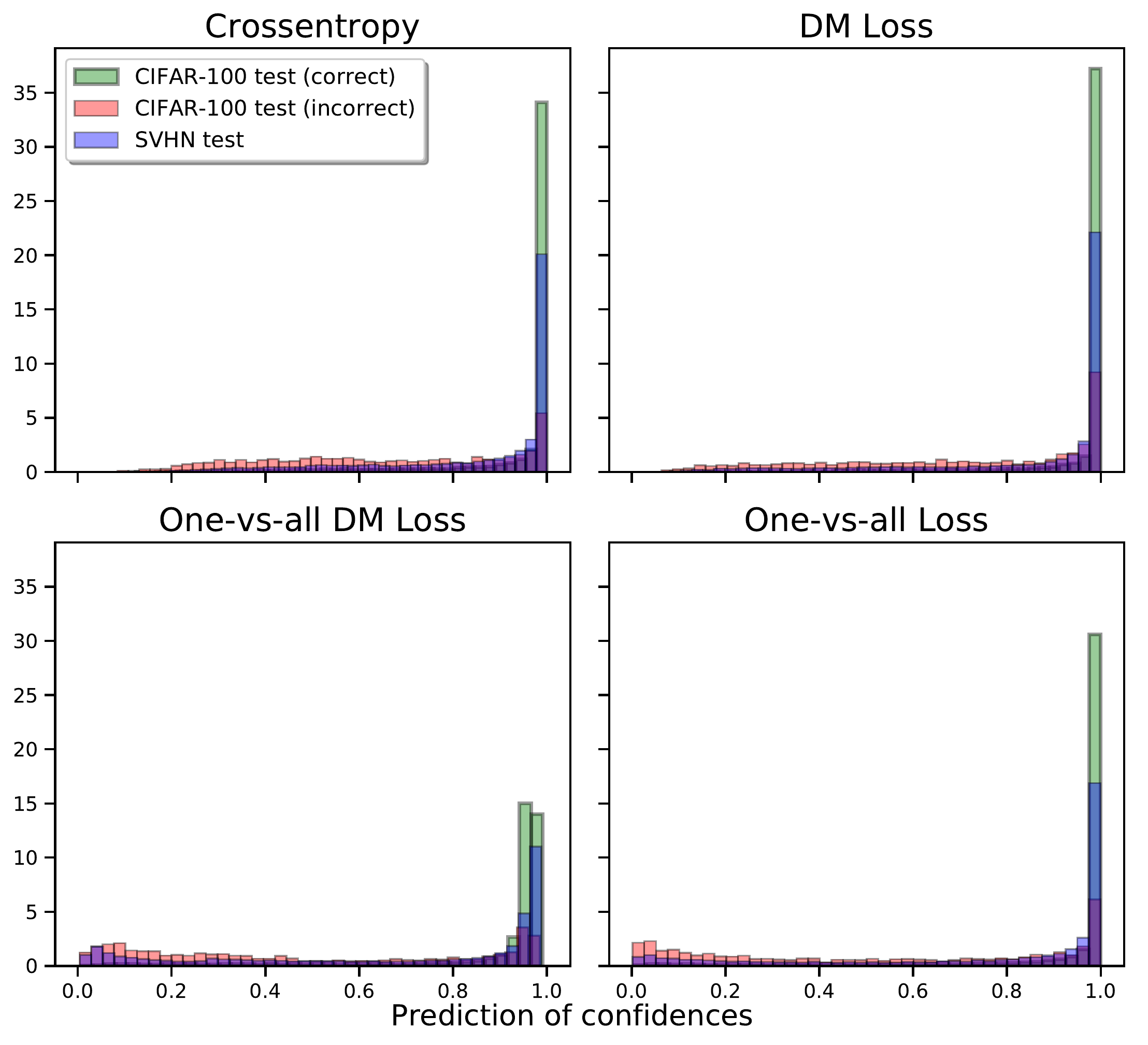}
         \reducespacebetweenfigureandcaption 
    } \end{subfigure}
\caption{%\textbf{
        Histograms of predicted confidences for a Wide ResNet 28-10 architecture trained with different loss functions on the CIFAR-100 dataset, and tested on the OOD CIFAR-10 and SVHN datasets. We can see that there is a much higher overlap in the confidence distributions of incorrectly classified in-distribution test points and OOD test points, which results in poorer performance when using a scalar metric such as AUROC and AUPRC that aggregate over all thresholds in $[0, 1]$.}
        \label{fig:ood_conf_cifar100_svhn}
\end{figure*}

The disparity above can be explained by the fact that confidence versus accuracy curves measure the ability of a method to separate set 1 from sets 2+3, whereas AUROC and AUPRC measure the ability of a method to separate sets 1+2 (in-distribution) from set 3 (OOD). 
Specifically, the confidence versus accuracy curve measures accuracy only on the unrejected inputs, hence it does not penalize a method for assigning the same low confidence to incorrectly classified in-distribution test points and OOD test inputs. On the other hand, AUROC measures the ability to separate in-distribution from OOD inputs, so it does not penalize a method for assigning the same confidence to correctly and incorrectly classified in-distribution inputs. 

From Figure~\ref{fig:ood_conf_cifar100_svhn}, we can see that one-vs-all DM loss outputs much lower confidences for incorrectly classified in-distribution test points, while also outputting low confidences for OOD test points. Due to the overlap in the confidence histograms for these two subsets of data, a scalar metric such as AUROC and AUPRC will report lower performance, due to difficulty in distinguishing between the two at low confidences.

\begin{table}[ht]
\centering
\resizebox{0.48\textwidth}{!}{%
\begin{tabular}{c|l|cc}
\multirow{2}{*}{\textbf{$\mathcal{D}_{\text{in}}$}} &
  \multicolumn{1}{c|}{\multirow{2}{*}{\textbf{Loss function}}} &
  \multicolumn{2}{c}{\textbf{AUROC $(\uparrow)$ / AUPRC $(\uparrow)$}} \\ \cline{3-4} 
 &
  \multicolumn{1}{c|}{} &
  \multicolumn{1}{c|}{\textbf{$\mathcal{D}_{\text{ood}}$ = CIFAR-100}} &
  \textbf{$\mathcal{D}_{\text{ood}}$ = SVHN} \\ \hline
\multirow{4}{*}{CIFAR-10} & Crossentropy       & \multicolumn{1}{c|}{0.7334 / 0.6789}          & 0.7026 / 0.8058 \\
                          & DM Loss            & \multicolumn{1}{c|}{0.7478 / 0.7104}          & 0.8264 / 0.8940 \\
                          & One-vs-all Loss    & \multicolumn{1}{c|}{\textbf{0.8449 / 0.8037}} & \textbf{0.8934 / 0.9248} \\
                          & One-vs-all DM Loss & \multicolumn{1}{c|}{0.6805 / 0.6486}          & 0.6191 / 0.7725
\end{tabular}%
}
\caption{AUROC and AUPRC scores for OOD detection tasks for a Wide ResNet 28-10 model trained on the CIFAR-10 dataset and evaluated on the CIFAR-100 and SVHN OOD datasets.}
\label{tab:ood_cifar10}
\end{table}

\begin{table}[ht]
\centering
\resizebox{0.48\textwidth}{!}{%
\begin{tabular}{c|l|cc}
\multirow{2}{*}{\textbf{$\mathcal{D}_{\text{in}}$}} &
  \multicolumn{1}{c|}{\multirow{2}{*}{\textbf{Loss function}}} &
  \multicolumn{2}{c}{\textbf{AUROC $(\uparrow)$ / AUPRC $(\uparrow)$}} \\ \cline{3-4} 
 &
  \multicolumn{1}{c|}{} &
  \multicolumn{1}{c|}{\textbf{$\mathcal{D}_{\text{ood}}$ = CIFAR-10}} &
  \textbf{$\mathcal{D}_{\text{ood}}$ = SVHN} \\ \hline
\multirow{4}{*}{CIFAR-100} & Crossentropy       & \multicolumn{1}{c|}{0.8003 / 0.7587}          & 0.6420 / 0.7779          \\
                           & DM Loss            & \multicolumn{1}{c|}{0.7862 / 0.7488}          & \textbf{0.6744 / 0.8047} \\
                           & One-vs-all Loss    & \multicolumn{1}{c|}{\textbf{0.8019 / 0.7601}} & 0.6700 / 0.8028          \\
                           & One-vs-all DM Loss & \multicolumn{1}{c|}{0.7669 / 0.7274}          & 0.5966 / 0.7948         
\end{tabular}%
}
\caption{AUROC and AUPRC scores for OOD detection tasks for a Wide ResNet 28-10 model trained on the CIFAR-100 dataset and evaluated on the CIFAR-10 and SVHN OOD datasets.}
\label{tab:ood_cifar100}
\end{table}

 \section{Implementation Details}
 \subsection{2D Toy Dataset}
 \label{subsection:2d}
 The synthetic 2-dimensional toy dataset is generated by drawing 1000 points per class from a 2-dimensional normal distribution, where the points for the $j$th class are given by $\bm{X}_j \sim \mathcal{N}_2\left( \left( 20 \cos(\frac{j}{10 \times 2\pi}), 20 \sin(\frac{j}{10 \times 2\pi}) \right), 2\bm{I}\right)$. For training, we use a fully-connected neural network with 2 hidden layers containing 16 units each, trained on a batch size of 128 using SGD for 10000 steps. For all examples, the trained models achieve a training accuracy of 100\%.

 \subsection{CIFAR-10}
 \label{subsection:cifar10}
 We use the Tensorflow Datasets \citep{TFDS} implementations of both the CIFAR-10 \citep{Krizhevsky09learningmultiple} and CIFAR-10-C \citep{hendrycks2018benchmarking} datasets. For training, the CIFAR-10 dataset is augmented by padding with zeros, followed by random cropping, random flipping, and rescaling to $[-1, 1]$.
 We train the Resnet-20 v1 architecture on a batch size of 512 using the Adam optimizer with a learning rate schedule, and tune the following hyperparameters using random search with 100 trials within ranges specified as follows: learning rate (\num{1e-3} to \num{1}), momentum (\num{0.85} to \num{0.99}), and Adam's $\epsilon$ (\num{1e-8} to \num{1e-5}).
 
 \subsection{CIFAR-100}
 \label{subsection:cifar100}
 We use the Tensorflow Datasets implementation of CIFAR-100 \cite{Krizhevsky09learningmultiple} and generate the CIFAR-100-C dataset by calculating the 17 types of corruptions (with the exception of \texttt{motion\_blur} and \texttt{snow}) over 5 levels of intensity as defined in \citet{hendrycks2018benchmarking}. The CIFAR-100 dataset is augmented in a similar manner as CIFAR-10 during training, and we use a Wide ResNet 28-10 architecture with L2 regularization \citep{zagoruyko2016wide} trained on a batch size of 512 for 200 epochs using SGD with Nesterov momentum \citep{tran2018simple}. We tune the following hyperparameters using random search with 50 trials with ranges specified as follows: learning rate (\num{1e-4}, \num{1e-1}) and L2 weighing factor (\num{1e-5}, \num{1e-3}).
 \subsection{ImageNet}
 \label{subsection:imagenet}
 We use the Tensorflow Datasets implementation of the ImageNet \citep{ILSVRC15} and ImageNet-C \citep{hendrycks2018benchmarking} datasets. During training, the ImageNet dataset is preprocessed using the data augmentation pipeline commonly used by the Inception architecture \citep{shorten2019survey}. We use a ResNet-50 "v1.5" architecture \footnote{The v1.5 architecture differs slightly from the v1 architecture more commonly defined in \citet{he2016deep}. For further details, please see \url{http://torch.ch/blog/2016/02/04/resnets.html}.} with L2 regularization trained on a batch size of 4096 for 90 epochs using SGD with Nesterov momentum and a warmup learning rate schedule \citep{you2017large}. We tune the following hyperparameters using random search with 50 trials with ranges specified as follows: learning rate (\num{5e-1}, \num{1e1}) and L2 weighing factor (\num{1e-5}, \num{1e-3}). 
 \subsection{Intent Classification}
 \label{subsection:intent}
 We use the CLINC out-of-scope (OOS) intent classification data as defined in \citet{larson2019evaluation}, where the BERT tokenizer is used to pre-tokenize sentences with a maximum sequence length of 32. For training, we use the \texttt{BERT-Mini} architecture \citep{devlin2018bert, tsai2019small}, which is a 4 layer transformer with 256 hidden units per layer and 4 attention heads, trained on a batch size of 256 for 1000 epochs using the Adam optimizer, where the learning rate is tuned for 30 trials sampled uniformly between (\num{1e-5}, \num{1e-3}).

\end{document}